\documentclass[11pt, a4paper, logo]{gdmstyle/googledeepmind}

\pdftrailerid{redacted}

\makeatletter
\renewcommand\bibentry[1]{\nocite{#1}{\frenchspacing\@nameuse{BR@r@#1\@extra@b@citeb}}}
\makeatother

\newcommand*{\eg}{e.g.,\@\xspace}

\newcommand*{\ie}{i.e.,\@\xspace}

\let\originalleft\left
\let\originalright\right
\renewcommand{\left}{\mathopen{}\mathclose\bgroup\originalleft}
\renewcommand{\right}{\aftergroup\egroup\originalright}

\def\eqref#1{eq.~\ref{#1}}

\DeclareMathAlphabet{\mathsfit}{\encodingdefault}{\sfdefault}{m}{sl}
\SetMathAlphabet{\mathsfit}{bold}{\encodingdefault}{\sfdefault}{bx}{n}

\newcommand{\KL}{$\mathrm{KL}$\@\xspace}

\usepackage{algorithm}
\usepackage[noend]{algpseudocode}
\newcommand{\algrule}[1][.2pt]

\algnewcommand\algorithmicinput{\textbf{Input:}}
\algnewcommand\Input{\item[\algorithmicinput]}
\algnewcommand\algorithmicoutput{\textbf{Output:}}
\algnewcommand\Output{\item[\algorithmicoutput]}
\algnewcommand\algorithmicempty{~}
\algnewcommand\Empty{\item[\algorithmicempty]}

\usepackage{mathtools}

\usepackage{kantlipsum, lipsum}
\usepackage{dsfont}
\usepackage{gdmstyle/gdm-colors}
\usepackage{bm}      

\usepackage{ifthen}
\newboolean{isanon}
\setboolean{isanon}{false}

\newboolean{islong}
\setboolean{islong}{true}
\usepackage[dvipsnames]{xcolor}
\usepackage{xurl}
\usepackage{graphicx}
\usepackage{amsmath}

\usepackage{pifont}%
\usepackage[authoryear, round]{natbib}
\usepackage[pagebackref=true,colorlinks=true,allcolors=colorgreenfull]{hyperref}
\usepackage{cleveref}
\usepackage{scalerel}
\usepackage{caption}
\usepackage{wrapfig}
\usepackage{subcaption}
\usepackage{bbding}
\usepackage{placeins}
\usepackage{microtype}
\usepackage{csquotes}
\usepackage{enumitem}
\usepackage{xspace}
\usepackage{booktabs}
\usepackage{mathtools}
\usepackage{amsthm}

\usepackage{tikz}
\usetikzlibrary{calc}
\usetikzlibrary{angles,quotes} 

\usepackage[super]{nth}
\usepackage{multirow}
\usepackage{rotating}
\usepackage{amssymb}
\usepackage[export]{adjustbox}
\usepackage{subcaption}
\usepackage{tcolorbox}
\everypar{\looseness=-1}

\usepackage{xspace}
\usepackage{algorithm}
\usepackage{algpseudocode}
\newcommand{\piref}{{\pi_\text{ref}}}
\newcommand{\piBoN}{{\pi_\text{BoN}}}
\newcommand{\pl}{{p_<}}
\newcommand{\ple}{{p_\leq}}

\newtheorem{theorem}{Theorem}

\newcommand{\SFT}{SFT\@\xspace}
\newcommand{\RLHF}{RLHF\@\xspace}
\newcommand{\LLM}{LLM\@\xspace}
\newcommand{\LLMs}{LLMs\@\xspace}
\newcommand{\GEMMA}{Gemma\@\xspace}
\newcommand{\BOND}{\texttt{BOND}\@\xspace}
\newcommand{\JBOND}{\texttt{J-BOND}\@\xspace}
\newcommand{\BON}{Best-of-\texttt{N}\@\xspace}
\newcommand{\BONS}{Best-of-\texttt{N}$^2$\@\xspace}
\newcommand{\BOn}{Best-of-\texttt{n}\@\xspace}

\renewcommand*{\backrefalt}[4]{%
    \ifcase #1 \footnotesize{(Not cited.)}%
    \or        \footnotesize{(p.~#2)}%
    \else      \footnotesize{(pp.~#2)}%
    \fi}

\title{\BOND: Aligning LLMs with \BON Distillation}

\correspondingauthor{piergs@google.com}

\keywords{LLM, Alignment, RLHF, \BON}

\author{Pier~Giuseppe~Sessa}
\author{Robert~Dadashi}
\author{Léonard~Hussenot}
\author{Johan~Ferret}
\author{Nino~Vieillard}
\author{Alexandre~Ramé}
\author{Bobak~Shariari}
\author{Sarah~Perrin}
\author{Abe~Friesen}
\author{Geoffrey~Cideron}
\author{Sertan~Girgin}
\author{Piotr~Stanczyk}
\author{Andrea~Michi}
\author{Danila~Sinopalnikov}
\author{Sabela~Ramos}
\author{Amélie~Héliou}
\author{Aliaksei~Severyn}
\author{Matt~Hoffman}
\author{Nikola~Momchev}
\author{Olivier~Bachem}

\affil{Google DeepMind}

\begin{abstract}
Reinforcement learning from human feedback (RLHF) is a key driver of quality and safety in state-of-the-art large language models.
Yet, a surprisingly simple and strong inference-time strategy is \BON sampling that selects the best generation among $N$ candidates.
In this paper, we propose \BON Distillation (\BOND), a novel RLHF algorithm that seeks to emulate \BON but without its significant computational overhead at inference time. Specifically, \BOND is a distribution matching algorithm that forces the distribution of generations from the policy to get closer to the \BON distribution. We use the Jeffreys divergence (a linear combination of forward and backward \KL) to balance between mode-covering and mode-seeking behavior, and derive an iterative formulation that utilizes a moving anchor for efficiency.
We demonstrate the effectiveness of our approach and several design choices through experiments on abstractive summarization and \GEMMA models.
Aligning \GEMMA policies with \BOND outperforms other RLHF algorithms by improving results on several benchmarks.
\end{abstract}

\begin{document}

\maketitle

\section{Introduction}
\label{sec:introduction}

State-of-the-art large language models (\LLMs) such as Gemini \citep{gemini2023,reid2024gemini} and GPT-4 \citep{openai2023gpt4} are generally trained in three stages. First, \LLMs are pre-trained on large corpora of knowledge using next-token prediction \citep{radford2018gen,radford2019language}. Second, the pre-trained models are fine-tuned to follow instructions via supervised fine-tuning (\SFT) \citep{JMLR:v21:20-074,wei2022finetuned}. Lastly, reinforcement learning from human feedback (\RLHF) \citep{christiano2017deep, ziegler2019fine, stiennon2020learning} is used to further increase the quality of generations. The \RLHF step generally consists of learning a reward model (RM) \citep{ouyang2022training} on human preferences and then optimizing the \LLM to maximize predicted~rewards using reinforcement learning algorithms.

\textbf{RLHF algorithms and their challenges.}
Fine-tuning LLMs with reinforcement learning (RL) is challenging \citep{casper2023open}, notably since it can cause \emph{forgetting}~\citep{french1992semi} of pre-trained knowledge, and since loopholes in the RM \citep{faultyreward2016,pan2022the} can cause \emph{reward hacking} \citep{askell2021general, skalse2022defining}.
The standard strategy is to use policy-gradient methods~\citep{williams1992simple} with \KL regularization towards the SFT policy.
Those RL algorithms seek Pareto-optimal policies with high reward at low \KL, to preserve the general capabilities of the original model and tackle the misalignment \citep{ngo2022alignment} concerns.

\textbf{\BON sampling.}
In practice, a surprisingly simple inference-time approach is often used to improve the quality of generations: \BON sampling~\citep{stiennon2020learning}. It consists of drawing $N$ candidate generations from the reference (typically, supervised fine-tuned) model and selecting the one with the highest reward according to the RM.
This strategy empirically achieves excellent reward-\KL trade-offs~\citep{nakano2021webgpt,gao2022scaling,touvron2023llama2} but increases the computational cost by a factor of $N$.

\begin{figure}[t]
    \centering
    \includegraphics[width=\textwidth]{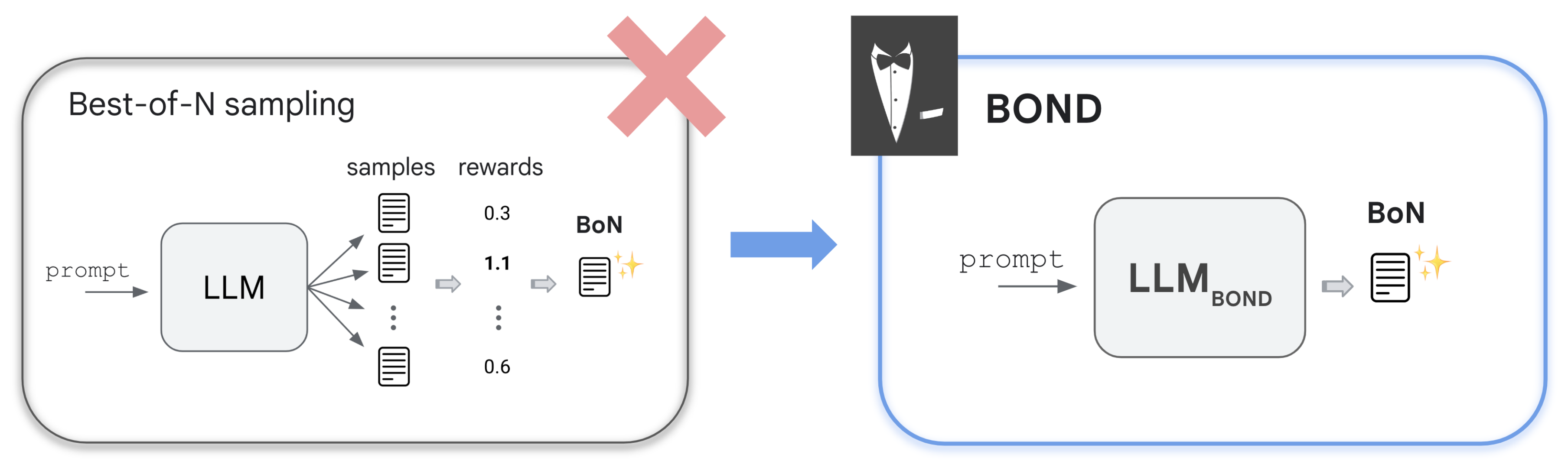}
    \caption{\BON is an \emph{inference-time} strategy that selects the best generation among $N$ candidates from a reference LLM policy, improving quality at the cost of a large computational (need to sample and score $N$ times from the model). In contrast, the proposed \BOND approach aims at obtaining a fine-tuned policy that can directly sample the \BON generation. This would inherit the quality of \BON sampling,  while requiring a single sample at inference time. We achieve this by distilling the \BON strategy into the policy via online \emph{distribution matching}.
    }
    \label{fig:bond_idea}
\end{figure}

\textbf{\BOND.}
In this paper, we propose \BOND (\BON Distillation), a novel RLHF algorithm that \emph{learns} a policy that achieves the strong performance of \BON sampling but, crucially, requires only a single sample at inference time, as depicted in~\Cref{fig:bond_idea}.
Our key idea is to cast the alignment of the policy as a distribution matching problem, where we fine-tune the policy to emulate the \BON distribution.
To achieve this, we first derive an analytical expression for the \BON distribution. This allows us to consider and optimize different divergence metrics. We first show how to minimize the \emph{forward} \KL divergence using samples from the \BON strategy, leading to a standard imitation learning setup with a mode covering behavior. We also show how to minimize the \emph{backward} \KL, leading to a new form of quantile-based advantage, which does not depend on the reward scale, and corresponds to a mode seeking behavior. Then, we propose to minimize a linear combination of forward and backward \KL, also known as \emph{Jeffreys divergence}, which retains the best of both approaches. Furthermore, to optimize performance while keeping a reduced sample-complexity, we propose an \emph{iterative} \BOND approach which consists of iteratively distilling the \BON of a moving anchor policy. Finally, based on the aforementioned ideas, we propose \JBOND (J for Jeffreys), a novel, stable, efficient and practical \RLHF algorithm to align LLMs.

\textbf{Experiments.}
We first demonstrate the effectiveness of \BOND and of our design choices on the abstractive summarization XSum~\citep{narayan-etal-2018-dont} task.
Then, in \Cref{sec:expe}, we apply \JBOND to align \GEMMA \citep{team2024gemma} policies.
\JBOND not only improves the \KL-reward Pareto front compared to standard RL algorithms, but also enhances performance on academic benchmarks and side-by-side comparisons against open-source variants such as Mixtral~\citep{jiang2024mixtral}.

\section{Problem Setup}
\label{sec:context}%

We consider a LLM based on the transformer \citep{vaswani2017transformer} architecture, defining a policy $\pi(x, \cdot)$ by auto-regressively generating token sequences $y$ from the prompt $x$.
Given a pre-trained and typically supervised fine-tuned reference policy $\piref$, we seek to further align it to human preferences. To achieve this, throughout the rest of the paper we assume access to a reward model (RM) which we denote as $r(\cdot)$, trained to reflect human preferences.

\textbf{Standard \RLHF.} Most RL algorithms optimize a linear combination of the expected reward and a \KL divergence between the current and reference policy: 
\begin{equation}
    \pi_\text{RL} = \operatorname{argmax}_{\pi} \mathbb{E}_{\pi}\left[r(y)\right] - \beta_\textrm{RL} \cdot  \operatorname{KL}\left(\pi\mid\mid \piref\right), \label{eq:RL_objective}
\end{equation}
with regularization strength $\beta_\textrm{RL}\geq 0$. This \KL regularization forces the policy to remain close to its initialization $\piref$ \citep{geist2019theory,lazaridou2020multi}, reducing forgetting~\citep{french1992semi} and reward hacking \citep{skalse2022defining}.
\Cref{eq:RL_objective} is usually optimized with online algorithms, as they perform better than their offline counterparts \citep{tang2024understanding}.
Moreover, simple methods have demonstrated the best results, \eg REINFORCE~\citep{williams1992simple} with a sampled baseline for variance reduction \citep{li2023remax,ahmadian2024back} outperform PPO~\citep{schulman2017proximal}.

\textbf{\BON.}
A complementary alignment strategy is \BON, which is an inference-time strategy that involves sampling multiple times from $\piref$ and selecting the generation with highest reward according to the RM $r$.
In contrast to \RLHF strategies, \BON does not fine-tune the weights of the LLM, but instead modifies the inference procedure.
\BON was empirically shown to be efficient~\citep{touvron2023llama2} when looking at reward/\KL trade-offs, and comes with theoretical guarantees~\citep{qiping2024asymptotics} in terms of Pareto-optimality.
Unfortunately, \BON comes at a significantly higher inference cost which increases linearly with $N$, since producing $N$ generations is (in general) $N$ times more costly than sampling a single one.

Motivated by the above considerations, we propose a novel alignment method which we name \BOND for \BON Distillation. 
The goal of \BOND is to \emph{distill the \BON strategy into the policy}.
This allows the policy to reach the strong performance of \BON sampling, while requiring only \emph{a single sample} at inference time. We outline our overall approach in the next section.

\section{The \BOND Approach}
We formulate the \BOND approach in two main steps. First, we derive an analytical expression for the \BON distribution (\Cref{sec:BoN_distribution}). Second, using the derived expression, we phrase the problem as a \emph{distribution matching} problem (\Cref{sec:bond_objective}), \ie we want to steer the policy closer to the \BON distribution. 
In \Cref{sec:connection_with_rlhf}, we draw insightful connections between \BOND and standard \RLHF.
\subsection{The \BON distribution}%
\label{sec:BoN_distribution}%
In this section, we derive the exact analytical distribution of \BON sampling and study its properties. 
For simplicity, we drop the context $x$ from all notation without loss of generality and assume that the reward $r(y)$ induces a strict ordering on all generations $y$\footnote{To distinguish between generations with the same reward, ties can be broken with any arbitrary strict ordering.}. We can affirm the following main theorem (proof in~\Cref{sec:proof_thm}).
\begin{theorem}
\label{thm:bondist}
For any generation $y$, let
\begin{equation}
    \pl(y) = \mathbb{P}_{y'\sim\piref}\left[r(y') < r(y)\right]
\end{equation}
denote the probability that a random generation $y'$ from $\piref$ is strictly worse than $y$ 
and let
\begin{equation}
    \ple(y) = \mathbb{P}_{y'\sim\piref}\left[r(y') \le r(y)\right],
\end{equation}
the probability that $y'$ is not better than $y$ (thus including the equality case).
Then, the probability that $y$ is the output of \BON sampling is given by
\begin{equation}
    \piBoN(y) 
    = \piref(y) \times \underbrace{\ple(y)^{N-1}}_{\mathtt{(A)}} \times \underbrace{\sum_{i=1}^N \left[\frac{\pl(y)}{\ple(y)}\right]^{i-1}}_{\mathtt{(B)}}.
    \label{eq:bonproba}
\end{equation}
\end{theorem}
\textbf{Interpretation.}
\Cref{thm:bondist} provides an intuitive explanation on the behavior of \BON sampling: it essentially reweights the original sampling distribution $\piref$, by the multiplicative terms $\mathtt{(A)}$ and $\mathtt{(B)}$.\looseness=-2

The term $\mathtt{(A)}$ corresponds to a penalty exponential in $N$ based on the fraction of generations (for the same prompt) that are worse or equal to the considered generation $y$. 
Intuitively, this ensures that we sample exponentially less from bad generations when increasing $N$.

The term $\mathtt{(B)}$ is an additional correction factor due to the potential of collisions among generations.
Importantly, it is at most linear in $N$ as it is always bounded within $[1, N]$:
\begin{equation}
    1 \leq 1 + \sum_{i=2}^N \left[\frac{\pl(y)}{\ple(y)}\right]^{i-1} = \sum_{i=1}^N \left[\frac{\pl(y)}{\ple(y)}\right]^{i-1} \leq \sum_{i=1}^N 1 \leq N
\end{equation}
It achieves its minimum at $1$ for the worst generation $y_-$ since we have $\pl(y_-)=0$ by definition.
This is not surprising, as we need to sample $y_-$ exactly $N$ times in a row and which corresponds to $\piBoN(y_-) = \piref(y_-)^N$ (note that $\pl(y_-)=\piref(y_-)$).
In contrast, if the likelihood of individual generations $y$ are low and such generations are good, then $\pl(y)$ is almost $\ple(y)$ and term $\mathtt{(b)}$ is close to $N$.
Intuitively, this corresponds to the case where sampling a generation $y$ multiple times is unlikely. In the extreme case when $\piref$ is a continuous distribution, term $\mathtt{(B)}$ is constant and equal to $N$ (see~\Cref{app:continuous_bon}). \looseness=-1

\subsection{The \BOND objective}
\label{sec:bond_objective}

The analytical characterization of the \BON distribution allows us to formulate \BOND as a distribution matching problem. That is, we want to solve the objective:
\begin{equation}
\label{eq:bond_objective}
\pi_\texttt{BOND}= \arg\min_{\pi\in \Pi} \, D(\pi \, \| \, \piBoN),
\end{equation}
where $D(\cdot \, \| \, \cdot)$ is a divergence metric steering the training policy $\pi$ towards $\piBoN$.
For this, a toolbox of possible divergences exist in the literature including, \eg forward and backward \KL~\citep{Kullback59}. Moreover, we can employ existing distribution matching techniques to estimate $D$ from online and offline samples. We defer the choice of suitable divergences and resulting \BOND algorithms to \Cref{sec:bond_algorithms}.\looseness=-1

\subsection{Connection with standard \RLHF}\label{sec:connection_with_rlhf}

In this section, we draw important connections between the two seemingly different objectives of standard RLHF (\Cref{eq:RL_objective}) and \texttt{BOND} (\Cref{eq:bond_objective}).

It is well known (see, \eg~\cite{vieillard2020leverage, rafailov2023direct}) that the policy maximizing the RLHF objective from \Cref{eq:RL_objective} is:
\begin{equation}\label{eq:RL_solution}
    \pi_\text{RL}(y) \propto \piref(y) \exp\left(\frac1\beta_\textrm{RL} r(y)\right).
\end{equation}
From the derived expression of $\piBoN$ in \Cref{thm:bondist}, we see that the \BON sampling distribution coincides with the optimal solution of standard RLHF when using the following specific \BOND reward:
\begin{equation}
    r_\texttt{BOND}(y) = \underbrace{\log \ple(y)}_{\texttt{(A)}} + \underbrace{\frac1{N-1} \log\sum_{i=1}^N \left[\frac{\pl(y)}{\ple(y)}\right]^{i-1}}_{\texttt{(B)}},
    \label{eqn:rlreward}
\end{equation}
and the specific regularization strength $\beta_\text{BOND} = \frac1{N-1}$.
The term $\texttt{(B)}$ corresponds to the correction factor in \Cref{thm:bondist},  which is bounded in $\left[0, \frac{\log N}{N-1}\right]$ for all generations $y$. Instead term $\texttt{(A)}$ lies in $(-\infty, 0]$. 
This provides two interesting insights for \BON sampling:
\begin{enumerate}
    \item \BON sampling corresponds to the solution of a standard \KL-regularized RLHF problem where the choice of $N$ determines the level of KL regularization.
    \item \BON sampling corresponds to optimizing the expected log reward quantile, \ie the log likelihood that the generation has larger reward than a random sample from the reference distribution.
    Interestingly, due to the concavity of the logarithm, $r_\texttt{BOND}(y)$ strongly encourages the model to avoid bad generations rather than encouraging to generate good ones.
    Moreover, $r_\texttt{BOND}(y)$ is \emph{invariant to monotone transformations of the reward} $r(\cdot)$, since it depends only on the rank among the generations.
    We conjecture that both these features make the \BOND reward $r_\texttt{BOND}(y)$ more robust to reward hacking compared to standard RLHF.
\end{enumerate}

The connection to RLHF also inspires the proposed approach in this manuscript: 
if we can compute the \BOND reward or equivalently the \BON distribution $\piBoN$, then we can steer the policy towards \BON via distribution matching.
In the next section we explore different algorithms to tackle the main underlying challenges.

\section{\BOND Challenges and Algorithms}\label{sec:bond_algorithms}
Implementing the \BOND approach induces the three following challenges: \emph{(1)} how to estimate the reward quantiles, \emph{(2)} which is the appropriate divergence metric to use, and \emph{(3)} how to choose the hyperparameter $N$ representing the number of sampled generations in \BON. We discuss and address these challenges in the next three subsections.

\subsection{Monte-Carlo quantile estimation} 
\label{sec:montecarlo}
One key difficulty in estimating the $\piBoN$ distribution is that we need to estimate the quantile
\begin{equation}
    \ple(y) = \mathbb{P}_{y'\sim\piref}\left[r(y') \leq r(y)\right],
\end{equation}
of a given generation $y$. The quantile $\ple(y)$ measures the quality of $y$ compared to generations from $\piref$ when conditioned on the same prompt (recall that we have suppressed the conditioning on $x$ in our notation).
A very simple but effective quantile estimation method is \emph{Monte-Carlo sampling}, sampling $k$ generations from $\piref$ and obtaining the following empirical estimate:
\begin{equation}
    \hat{p}_\leq(y) = \frac{1}{k} \sum_{i=1}^k \mathbb{I}\{ r(y_i) \leq r(y)\}.
\end{equation}
We found this to be a very effective in our experiments, even with a limited number of samples. 
In principle, though, one could also use alternative approaches, \eg training a learned quantile model (as we explore in \Cref{app:learned_quantiles}).

\subsection{Jeffreys divergence as a robust objective}\label{sec:jeffreys}

The choice of the divergence metric used in \BOND is of crucial importance: different divergences can steer the policy to very different solutions. Here, we propose the \emph{Jeffreys divergence} as a robust distribution matching objective.

The Jeffreys divergence~\citep{Jeffreys46} between two distributions is defined as:
\begin{equation}
    J_\text{effreys}^\beta(p \, \| \, q) :=  (1-\beta) \cdot  \underbrace{\text{KL}(q \, \| \, p)}_{\text{Forward KL}} \, + \, \beta \cdot \underbrace{\text{KL}(p \, \| \, q)}_{\text{Backward KL}}. \label{eq:jeffreys_div}
\end{equation}
The (generalized) Jeffreys divergence is a weighted average (with weight $\beta \in [0,1]$) between the forward and backward \KL divergence. 
Notably, when fine-tuning policy $p$, the forward $\text{KL}(q \, \| \, p)$ encourages that generations likely under $q$ are also likely under $p$, thus encouraging a \emph{mode-covering} behavior. Instead, the reverse $\text{KL}(p \, \| \, q)$ is well-known to have a \emph{mode-seeking} effect, steering policy $p$ to produce generations that have a high likelihood according to $q$~\citep{agarwal2024onpolicy}. While the forward \KL may produce over-spread distributions, the backward \KL can lead to policy and entropy collapses. Instead, we empirically show that the Jeffreys divergence inherits the best of both divergences, producing better aligned policies.

In the context of \BOND, this translates into minimizing the divergence $J_\text{effreys}^\beta(\pi \, \| \, \piBoN)$ which we can estimate using samples from the training policy $\pi$ and reference policy $\piref$ as follows.

\textbf{Estimation of the forward KL.} The forward KL defined as
\begin{equation}
    \text{KL}(\piBoN \, \| \, \pi) = \mathbb{E}_{y \sim \piBoN}\left[\log \piBoN(y) - \log \pi(y)\right]
\end{equation}
can be estimated directly drawing samples from the $\piBoN$ (\ie sampling $N$ times from $\piref$ and selecting the best one) and can be seen as a supervised fine-tuning loss on the \BON samples:
\begin{equation}
   \nabla_\pi \text{KL}(\piBoN \, \| \, \pi) = - \mathbb{E}_{y \sim \piBoN} \nabla \log \pi(y).
\end{equation}

\textbf{Estimation of the backward KL.} The backward KL defined as
\begin{equation}\label{eq:backward_kl_from_bon}
    \text{KL}(\pi \, \| \, \piBoN) = \mathbb{E}_{y \sim \pi}\left[\log \pi(y) - \log \piBoN(y) \right]
\end{equation}
can be estimated from the policy samples (note the expectation w.r.t. $\pi$) and their estimated log-likelihood under $\piBoN$. In particular, by the analogies drawn in \Cref{sec:connection_with_rlhf}, we show (in \Cref{sec:pg_equivalence}) that its gradient coincides with a policy gradient (\eg used by REINFORCE~\citep{williams1992simple} in standard \RLHF):
\begin{equation}
   \nabla_\pi \text{KL}(\pi \, \| \, \piBoN) =  - (N-1)  \mathop{E}_{y\sim \pi }\left[  \nabla_\pi \log \pi(y) \big( r_\texttt{BOND}(y) - \beta_\texttt{BOND} \left(\log \pi(y) - \log \piref(y) \big) \right) \right], \label{eq:backward_kl_policy_gradient}
\end{equation}
with the equivalent reward $r_\texttt{BOND}$ and regularization $\beta_\texttt{BOND}$ defined in \Cref{sec:connection_with_rlhf}. Note that $r_\texttt{BOND}(y)$ depends on the true unknown quantile $\ple(y)$ and on the correction factor \texttt{(B)} defined in~\Cref{eqn:rlreward}. In practice, we substitute the true quantile by its estimate, while we observed the correction factor does not play a significant role. Thus, we use $r_\texttt{BOND}(y) = \hat{p}_\leq(y)$. Moreover, to reduce the resulting variance, we use a policy gradient baseline~\citep{Sutton+Barto:1998} which we compute as the average return for the other generations in the batch.

Thus, the overall $J_\text{effreys}^\beta$ loss is a linear weighted combination between a supervised fine-tuning and a policy gradient loss.

\begin{figure}[t]
    \centering
    \includegraphics[width=0.32\textwidth]{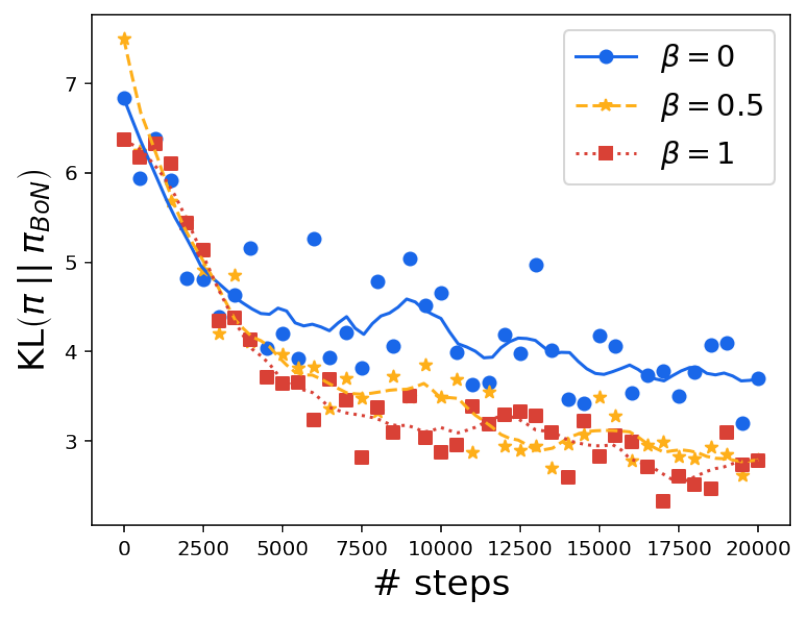}
    \includegraphics[width=0.325\textwidth]{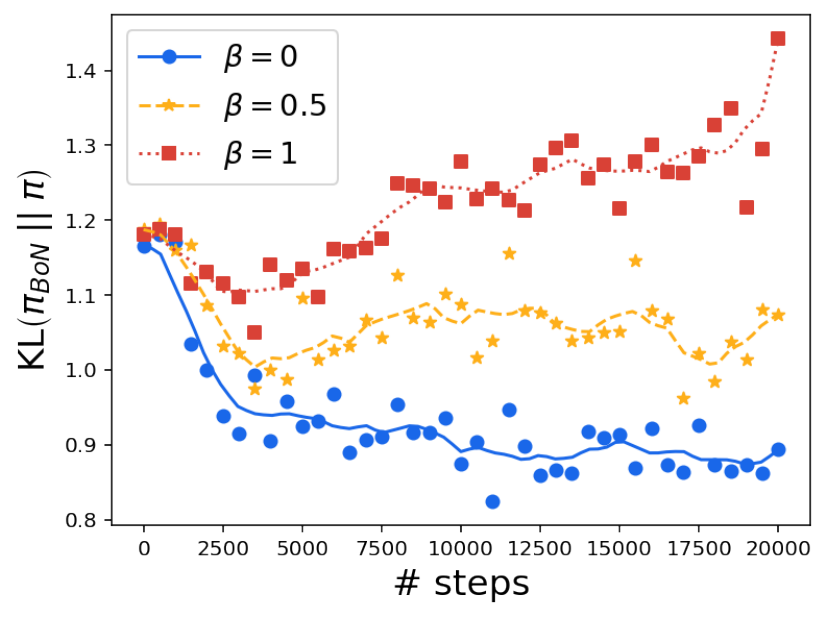}
    \includegraphics[width=0.33\textwidth]{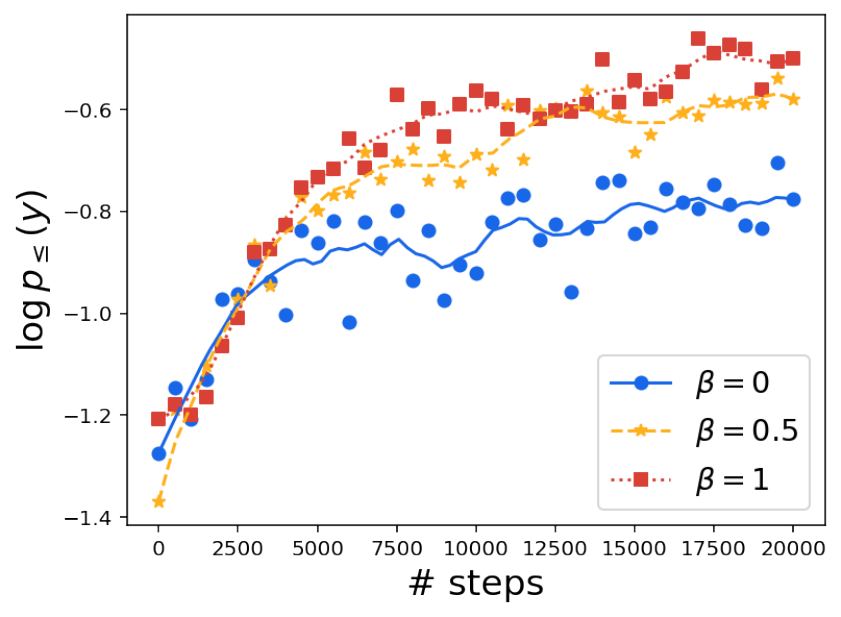}
    \caption{\BOND with $N=8$ and different values of $\beta$ for the Jeffreys divergence objective (\emph{cf.} \Cref{eq:jeffreys_div}). When using $\beta=0.5$ (Jeffreys divergence), \BOND minimizes both backward (left plot) and forward (middle plot) KL divergences from $\piBoN$, achieving best of both objectives ($\beta=0$ and $\beta=1$). Moreover, it optimizes the reward quantiles (right plot) significantly more than when using $\beta=0$ (forward KL minimization). Results for $N=4$ and $N=16$ are reported in \Cref{app:additional_results_jeffreys}.}
    \label{fig:jeffreys_ablation}
\end{figure}

\textbf{Experiments.} In \Cref{fig:jeffreys_ablation}, we consider the abstractive summarization XSum task~\citep{narayan-etal-2018-dont} with $\piref$ being a T5 supervised fine-tuned policy and $r(\cdot)$ being a T5 NLI reward model~\citep{roit2023factually}. We run \BOND with $J_\text{effreys}^\beta$ objective and $\beta \in \{0, 0.5, 1\}$. We use 16 MC samples (per prompt) to estimate quantiles during training. During eval (each 500 training steps), we use 32 MC samples to estimate the backward and forward KL divergences between the training policy and the $\piBoN$ distribution. We set $N=8$ in \Cref{fig:jeffreys_ablation}, and provide similar plots in \Cref{app:additional_results_jeffreys} for $N=4$ and $16$. 
The results confirm our intuition: the Jeffreys divergence ($\beta=0.5$) allows to minimize both divergences from $\piBoN$ (left and middle plot), compared to when solely the backward ($\beta=1$) or the forward ($\beta=0$) KL divergence is minimized. In addition, in the rightmost plot we report the reward log quantiles (averaged over the eval batches) of the training policy. Interestingly, \BOND with $\beta=0.5$ maximizes the quantiles similarly to the mode-seeking $\beta=1$ choice, while the mode-covering forward KL ($\beta=0$) lags behind.

\subsection{Iterative \BOND}\label{sec:iterative_bond}

Finally, we discuss the choice of the parameter $N$.
In practice, choosing $N$ may be difficult for three main reasons: \emph{(1)} As in standard RLHF, $N$ plays the role of regularization (see \Cref{sec:connection_with_rlhf}): a large $N$ improves downstream performance, but if $N$ is too large it will eventually cause reward over optimization~\citep{gao2022scaling}. \emph{(2)} The larger the $N$ the more the estimate of $\piBoN$ is sensitive to errors in the estimated quantiles (since $\piBoN(y) \propto p_\leq(y)^{N-1}$). \emph{(3)} Estimating the forward KL divergence requires sampling from $\piBoN$ which is prohibitive for large $N$.\looseness=-1

\begin{figure}[b]%
    \centering%
    \includegraphics[width=\textwidth]{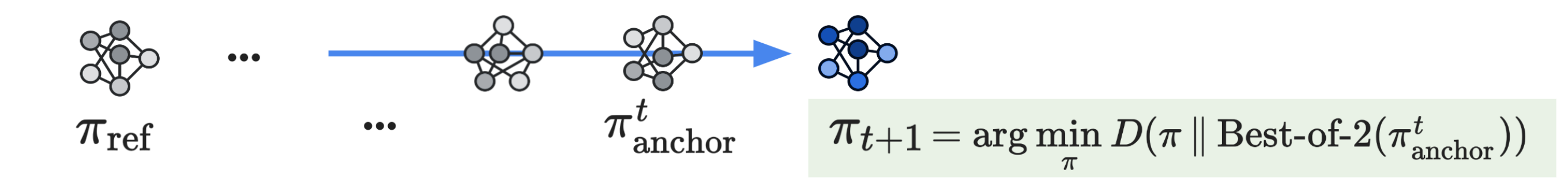}%
    \caption{Iterative \BOND approach. The policy $\pi_t$ is trained to iteratively distill a Best-of-$N$ (in the figure, $N=2$) version of a \emph{moving anchor}. This allows to continuously improve the policy performance without requiring to set a (large) $N$ upfront. Moreover, it leads to better training stability and a minimal computational complexity, since a small $n$ is used at each distillation step.}%
    \label{fig:iterative_bond}%
\end{figure}

To address the above challenges, we propose the \emph{iterative} \BOND approach. The approach is inspired by the fact that \BON sampling from a \BON distribution, coincides with \BONS sampling from the original distribution. More generally, by informally defining $\text{Bo$N$}(\cdot)$ as an operator that performs \BON sampling from a base distribution, we have:
\begin{equation}
    \underbrace{\text{Bo\texttt{N}}(\cdots\text{Bo\texttt{N}}(\text{Bo\texttt{N}}(}_{M \text{ times}}\piref))) \, \equiv \, \text{Bo\texttt{N}$^M$}(\piref).
\end{equation}
This suggests the key idea behind iterative \BOND: if we know how to distill the \BON distribution (\ie via \BOND), then we can apply \BOND recursively (say $M$ times), equivalently to distilling a Best-of-\texttt{N$^M$} of the initial distribution $\piref$. This allows fixing a small $n$ (\ie $n=2$) and running \BOND (with $N=n$) in an \emph{iterative fashion}, as an improvement operator. For this, we can introduce an auxiliary \emph{anchor policy} $\pi_\text{anchor}$ initialized as $\piref$. We can then run \BOND against $\pi_\text{anchor}$ (\ie we can distill the \BOn version of $\pi_\text{anchor}$) and, after a given number of distillation steps, update  $\pi_\text{anchor}$ to be the current training policy $\pi_t$.  
The overall approach is depicted in \Cref{fig:iterative_bond} and summarized in \Cref{alg:iterative-BOND}.\looseness=-1

In a nutshell, iterative \BOND allows exponential scaling to arbitrary large $N$ (in fact, it does not require setting $N$ in advance) while keeping a reduced sample complexity and a stable optimization. The claim is validated in the results below.

\begin{figure}[t]
    \centering
    \includegraphics[width=0.325\textwidth]{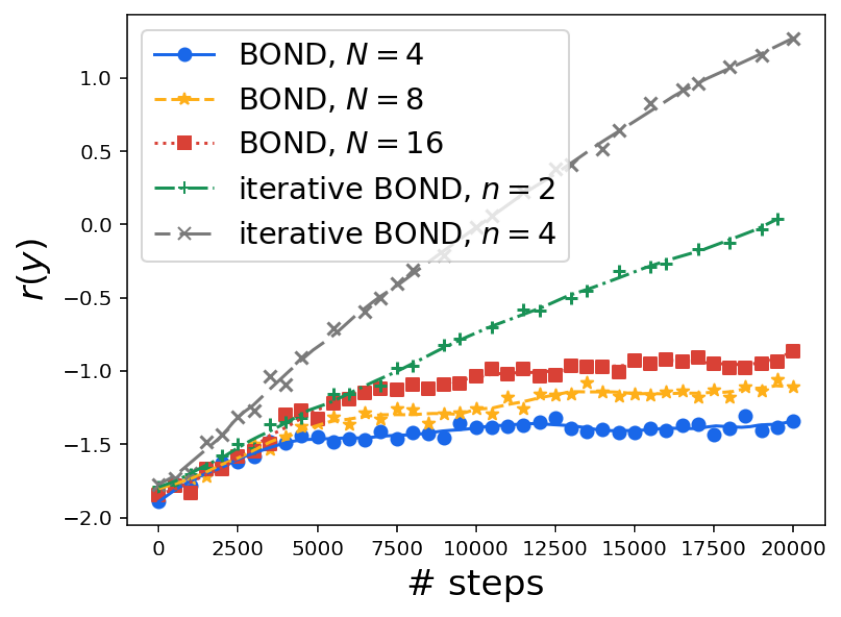}
    \includegraphics[width=0.325\textwidth]{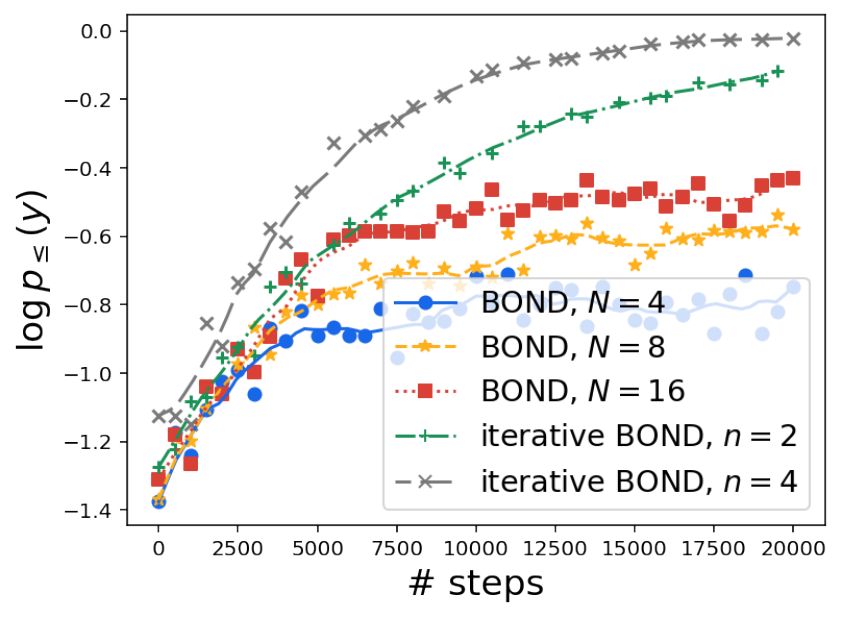}
    \includegraphics[width=0.325\textwidth]{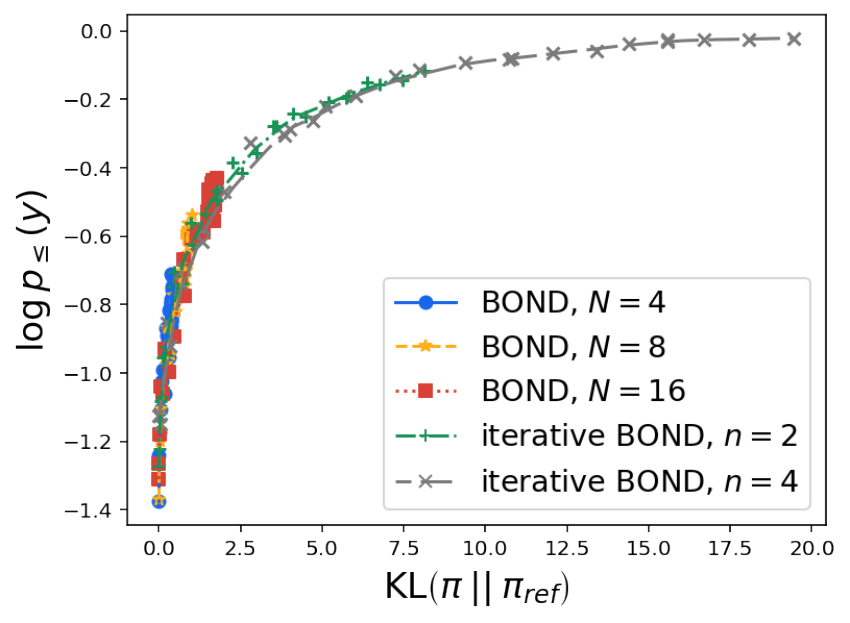}
    \caption{Iterative \BOND (with $n=2$ and $n=4$) compared to \BOND with $N=4,8,16$. Iterative \BOND continuously improves rewards (left plot) and log quantiles (middle plot), while they saturate for non-iterative \BOND (the smaller the $N$, the sooner). It allows achieving the same reward/\KL trade-off (right plot) as non-iterative \BOND while keeping a small $n$ and smoothly moving away from $\piref$.}
    \label{fig:iterative_bond_results}
\end{figure}

\textbf{Experiments.} In \Cref{fig:iterative_bond_results}, we consider the same experimental setup than in \Cref{fig:jeffreys_ablation} from \Cref{sec:jeffreys}, fix the \BOND objective to $J_\text{effreys}^{0.5}$ and run iterative \BOND with $n \in \{2, 4\}$ where the moving anchor is updated every 1000 steps. 
We report the average reward (left plot) and average log quantile (middle plot) obtained during training and compare them  to (non-iterative) \BOND run with $N\in \{4, 8, 16\}$. As expected, both reward signals saturate early for non-iterative \BOND (the smaller the $N$ the earlier the reward saturates), while the iterative \BOND approach continuously improves performance (the higher the $n$, the faster).
Moreover, in the rightmost plot we plot the obtained log quantiles against the \KL from the reference policy. The plot shows that iterative \BOND essentially has the same reward/\KL trade-off of the non-iterative \BOND runs, but crucially allows keeping a small $n$ and to smoothly but continuously move away from $\piref$. 

\begin{algorithm}[t]
\caption{Iterative \BOND (meta algorithm)}\label{alg:iterative-BOND}
\begin{algorithmic}
\State \textbf{Inputs:} $\piref$, $n \in \mathbb{N}$.
\State Initialize $\pi_0 = \piref$, $\pi_\text{anchor}^0 = \piref$.
\For{$ t = 0, \ldots, $} \hfill \text{\small \color{gray} */ iterations}
\State $\pi_{t+1} = \arg\min_{\pi\in \Pi} D(\pi \, \| \, \text{\BOn}(\pi_\text{anchor}^t))$ \hfill \text{\small \color{gray} */ distill the \BOn version of $\pi_\text{anchor}^t$}
\State $\pi_\text{anchor}^t = \pi_t$
\EndFor
\end{algorithmic}
\end{algorithm}

\section{The \JBOND Algorithm}
\label{sec:jbond}

In this section we present \JBOND, a concrete and practical \BOND algorithm motivated by the results discussed in the previous sections. We describe its main components below, and summarize it in the pseudo-code of \Cref{alg:jbond}.

\JBOND follows the template of iterative \BOND (\Cref{alg:iterative-BOND}) with $n=2$, \ie it fine-tunes policy $\pi_t$ to iteratively distill the Best-of-2 version of a moving anchor $\pi_\text{anchor}^t$, initialized as $\piref$. 
The name \JBOND stands for \emph{Jeffreys divergence} \BOND because it uses the Jeffreys divergence as distribution matching objective, \ie it minimizes $J_\text{effreys}^\beta(\pi \, \| \, \text{Best-of-2}(\pi_\text{anchor}^t))$ as defined in \Cref{sec:jeffreys}.

\textbf{Minimal sample complexity.} Compared to the \BOND algorithms tested in the previous section, \JBOND has a minimal sample complexity: for each prompt in the batch it generates $1$ sample from the policy $\pi_t$ and $2$ samples from the anchor $\pi_\text{anchor}^t$. While more anchor samples are generally useful for a better divergence estimation (in Section~\ref{sec:bond_algorithms} we used 16 MC samples), autoregressive sampling is the main bottleneck of online RLHF and we have thus opted for a practical approach working with a small number of samples.

\begin{algorithm}[t]
\caption{The \JBOND algorithm}\label{alg:jbond}
\begin{algorithmic}
\State \textbf{Inputs:}  Prompt dataset $\mathcal{D}$, reference policy $\piref$, reward $r(\cdot)$,  $\beta, \eta \in [0,1], \gamma \geq 0$.
\State Initialize policy and anchor $\pi_0 = \pi_\text{anchor}^0 = \piref$.
\For{$ t = 0, \ldots$}
 \State Sample batch of prompts $\mathcal{D}_t \subseteq \mathcal{D}$
 \State {For each $x\in \mathcal{D}_t$ generate:
 1 policy sample $y\sim \pi_t(x)$ and 2 anchor samples $y_1', y_2' \sim \pi_\text{anchor}^t(x)$}

 {\small \color{RoyalBlue}\texttt{*/ \textbf{Forward KL} \hfill */}}
 \State Extract Best-of-2 sample: $y'_\text{Bo2} = \arg\max_{y'\in \{y_1', y_2'\}} r(y')$.
 \State Compute forward KL gradient: \quad $G_\text{FW}(x,\pi_t) =  - \nabla_{\pi_t} \log \pi_t(x, y_\text{Bo2}')$

 {\small \color{RoyalBlue}\texttt{*/ \textbf{Backward KL} \hfill */}}
\State Compute $r_\texttt{J-BOND}(x,y)$ according to~\Cref{eq:jbond_reward}.
 \State Compute return: $R(x,y) = r_\texttt{J-BOND}(x,y) - (\log \pi_t(x,y) - \log \pi_\text{anchor}^t(x,y) )$.
 \State [Optional] Compute baseline $B$, e.g. average return of the other generations in the batch.
 \State Compute backward KL gradient: \quad 
   $G_\text{BW}(x,\pi_t) =  - \nabla_{\pi_t} \log \pi_t(x,y) \cdot \left( R(x,y) - B\right)$.

{\small \color{RoyalBlue}\texttt{*/ \textbf{Additional KL regularization} \hfill */}}
\State KL regularization gradient: \quad 
   $G_\text{Reg}(x,\pi_t) =  - \nabla_{\pi_t} \log \pi_t(x,y) \cdot ( \log \pi_t(x,y) - \log \pi_\text{anchor}^t(x,y))$.

 {\small \color{RoyalBlue}\texttt{*/ \textbf{Overall policy update: Jeffreys divergence + KL regularization} \hfill */}}

\State Update policy weights $\theta_{t+1}$ with the overall stochastic gradient: 
\vspace{-0.3em}
$$\mathbb{E}_{x\sim \mathcal{D}_t}\left[ (1-\beta)\cdot G_\text{FW}(x,\pi_t) + \beta \cdot G_\text{BW}(x,\pi_t)  + \gamma \cdot G_\text{Reg}(x,\pi_t) \right] $$

 {\small \color{RoyalBlue}\texttt{*/ \textbf{Update moving anchor} \hfill */}}
 \State Update anchor weights with EMA: \quad $
    \theta_\text{anchor}^{t+1} \leftarrow (1-\eta) \cdot \theta_\text{anchor}^t + \eta \cdot \theta_{t+1}$
\EndFor
\end{algorithmic}
\end{algorithm}

\textbf{Crude divergence estimate based on 2 anchor samples.} The policy and anchor samples are used to obtain a crude estimate of the forward and backward KL components of $J_\text{effreys}^\beta(\pi \, \| \, \text{Best-of-2}(\pi_\text{anchor}^t))$ as described next.

We can minimize the \emph{forward KL} as described in \Cref{sec:jeffreys}, by doing supervised fine-tuning on the best of the 2 anchor samples.
To minimize the \emph{backward KL}, we utilize the policy gradient-style loss of \Cref{eq:backward_kl_policy_gradient} replacing $r_\texttt{\BOND}(y)$ with a different reward which we denote as $r_\texttt{J-BOND}(y)$. The reason for this is that when only 2 anchor samples are available, the reward $r_\texttt{BOND}(y) = \log \hat{p}_\leq(y)$ would be quite uninformative due to $\hat{p}_\leq(y)$ being a very noisy MC estimate.
Let $y$ be the policy sample and $\{y_1', y_2'\}$ be the corresponding anchor samples, we instead define $r_\texttt{J-BOND}(y)$ as 
\begin{equation}\label{eq:jbond_reward}
   r_\texttt{J-BOND}(y) =  \begin{cases} -\log(16) & \text{if} \quad r(y) < \text{min}\{r(y_1'), r(y_2')\}\\ 0 & \text{otherwise} \end{cases}.
\end{equation}
That is, generation $y$ receives a negative reward of $-\log(16)$ if it has worse reward than \emph{both} anchor samples, while receives $0$ reward otherwise. The above definition is motivated by the following two main reasons: 
\begin{itemize}
    \item[(i)] We negatively reward $y$ only if it is worse than both the anchor samples, to mimic the concavity of the ideal (and unknown) reward function $r_\texttt{\BOND}=\log \ple(\cdot)$.
    \item[(ii)] We choose value $-\log(16)$ to ensure that: $\mathbb{E}_{y_1', y_2' \sim \pi_\text{anchor}^t}\big[  r_\texttt{J-BOND}(y)\big] = \log \ple(y)$ when $\ple(y)=0.5$.  The interested reader is referred to \Cref{sec:jbond_reward} for a derivation and illustration of this fact. In words, the value $-\log(16)$ calibrates the reward function $r_\texttt{J-BOND}(\cdot)$ so that, in expectation under the 2 anchor samples, it matches with the ideal reward $\log \ple(\cdot)$ for generations $y$ that have median reward (\ie when $\ple(y)=0.5$).
\end{itemize}

\textbf{Exponential Moving Average (EMA) anchor.} An important component of \JBOND, which refines the vanilla iterative \BOND of \Cref{sec:iterative_bond}, is the use of an \emph{Exponential Moving Average (EMA)} anchor. That is, instead of using a periodically updated anchor, we update the anchor weights $\theta_\text{anchor}^t$ \emph{at each fine-tuning step} as a moving average of the policy weights $\theta_t$:
\begin{equation}
\label{eq:ema_anchor}
    \theta_\text{anchor}^{t+1} \leftarrow (1-\eta) \cdot \theta_\text{anchor}^t + \eta \cdot \theta_{t+1}.
\end{equation}
Consistently with WARP \citep{rame2024warp}, we observed that this weight averaging procedure has a positive effect on training stability by reducing variance, and can improve the overall reward/KL trade-off of \JBOND. We provide an ablation in \Cref{sec:expe}.

\textbf{Additional KL regularization.} Finally, we further regularize the policy to stay closer to the moving anchor via an extra\footnote{Note that \KL regularization is already present in the backward \KL component of the Jeffreys divergence.} \KL regularization term, modulated by a tunable hyperparameter $\gamma \geq 0$. The scope is to further stabilize the policy updates, viewing the overall operator as a constrained optimization one:\looseness=-1
\begin{equation}
    \label{eq:j_bond_update_operator}
    \pi_{t+1} = \arg\min_{\pi \in \Pi} J_\text{effreys}^\beta(\pi \, \| \, \text{Best-of-2}(\pi_\text{anchor}^t)) \quad + \quad \gamma \cdot \text{KL}(\pi_t \, \| \, \pi_\text{anchor}^t ).
\end{equation}

\section{Experiments}
\label{sec:expe}

We test \JBOND on relevant use cases with the following main goals.
First, we ablate and showcase important aspects of \JBOND: the benefits of the EMA anchor, and the effects of the anchor speed and the additional KL regularization. Then, we compare \JBOND to classical RLHF baselines using REINFORCE, demonstrating its efficacy and better performance.

\begin{figure}[t]
    \centering
    \includegraphics[width=0.325\textwidth]{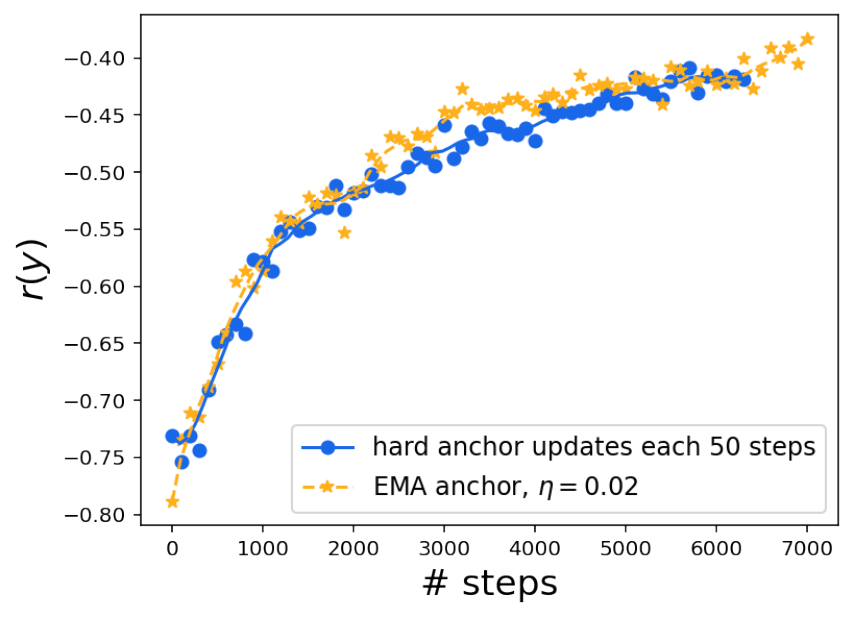}
    \includegraphics[width=0.325\textwidth]{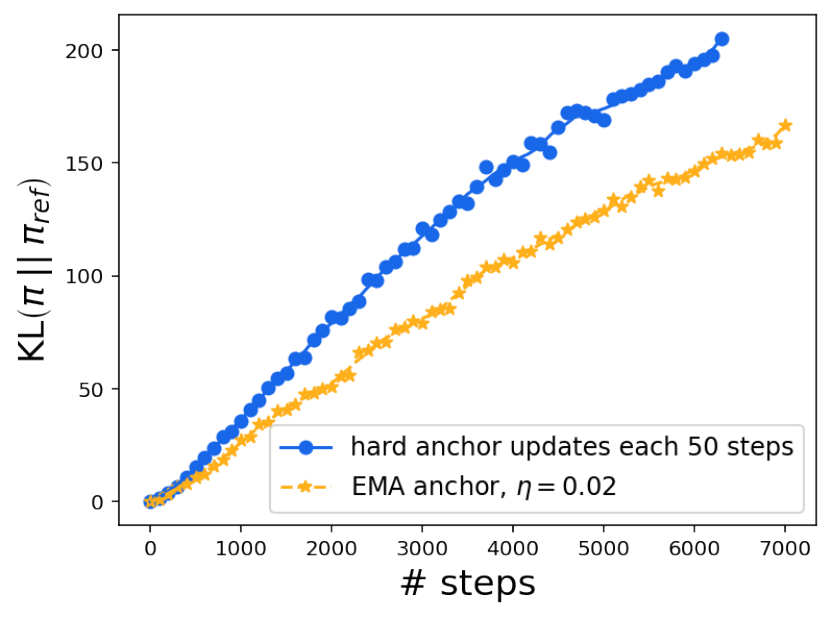}
    \includegraphics[width=0.325\textwidth]{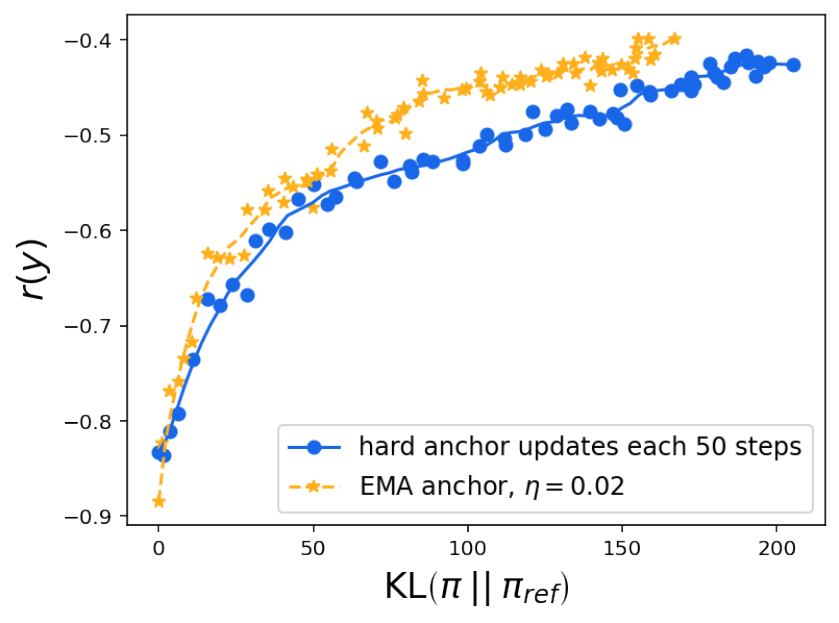}
    \caption{\JBOND with periodic anchor updates (every 50 steps) vs. EMA anchor ($\eta=0.02$) on Gemma 7B. While attaining the same reward (left), using the EMA anchor displays a significantly lower KL than the reference policy (middle) and thus a better reward/KL trade-off (right).}
    \label{fig:ema_benefit}
\end{figure}

\textbf{Setup.} We consider Gemma (2B and 7B) models~\citep{team2024gemma} which we aim to fine-tune into better conversational agents. For this task, we consider a set of conversational prompts $\mathcal{D}$, a reference policy $\piref$ previously supervised fine-tuned on similar prompts, and a large previously trained reward model $r(\cdot)$. We use a batch size 128 and the Adam optimizer~\citep{kingma2014adam} with learning rate $3e^{-6}$ and 100 warm-up steps. For the Jeffreys divergence objective, we set $\beta=0.5$.

\textbf{EMA vs. hard anchor updates.} We ablate the benefits of using an EMA moving anchor (\Cref{eq:ema_anchor}) compared to the periodically updated anchor used in Section~\ref{sec:iterative_bond}. For this, we run \JBOND with $\gamma=0$ and EMA coefficient $\eta = 0.02$ on Gemma 7B, and compare it with its variant where the anchor is only updated every $50$ steps. 
In \Cref{fig:ema_benefit}, we report the average reward of the policy during training (left plot), the KL from the reference policy $\piref$ (middle plot), and the resulting reward/KL trade-off for the two runs.
Both runs produce the same reward increase profile (this is not surprising since an EMA with $\eta=0.02$ roughly corresponds to an update period of 50 steps) but, crucially, \JBOND \emph{with an EMA anchor displays a significantly lower KL increase} and, as a result, a better reward/KL trade-off.

\begin{figure}[t]
    \centering
    \includegraphics[width=0.33\textwidth]{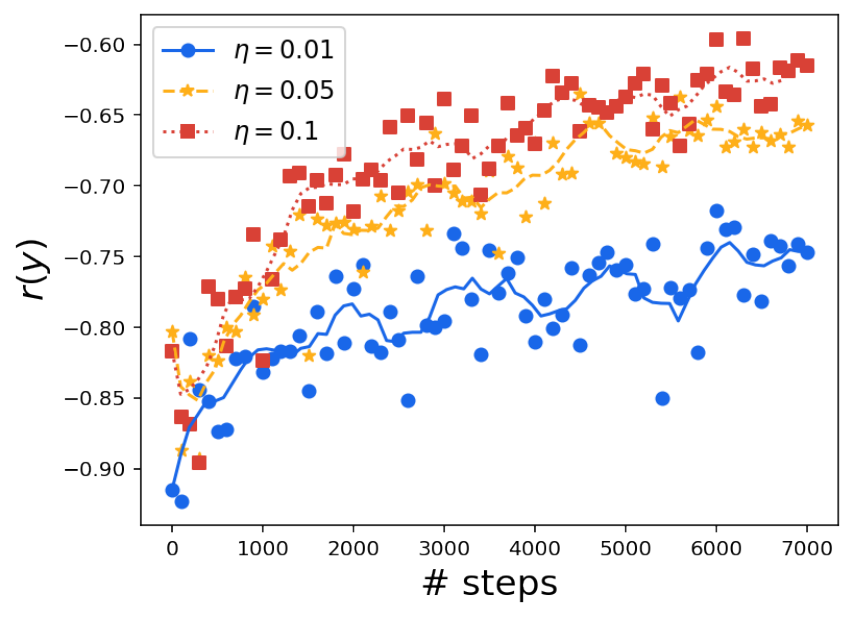}
    \includegraphics[width=0.32\textwidth]{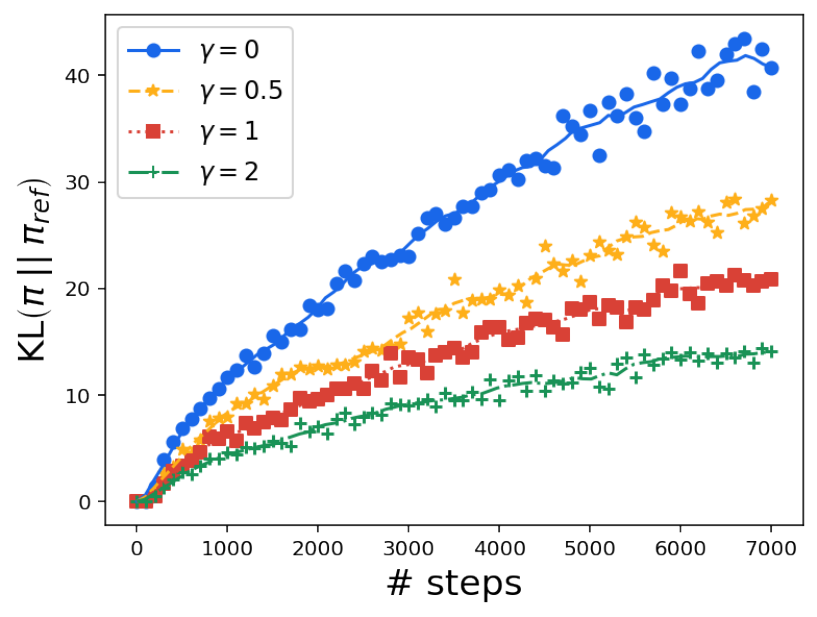}
    \includegraphics[width=0.33\textwidth]{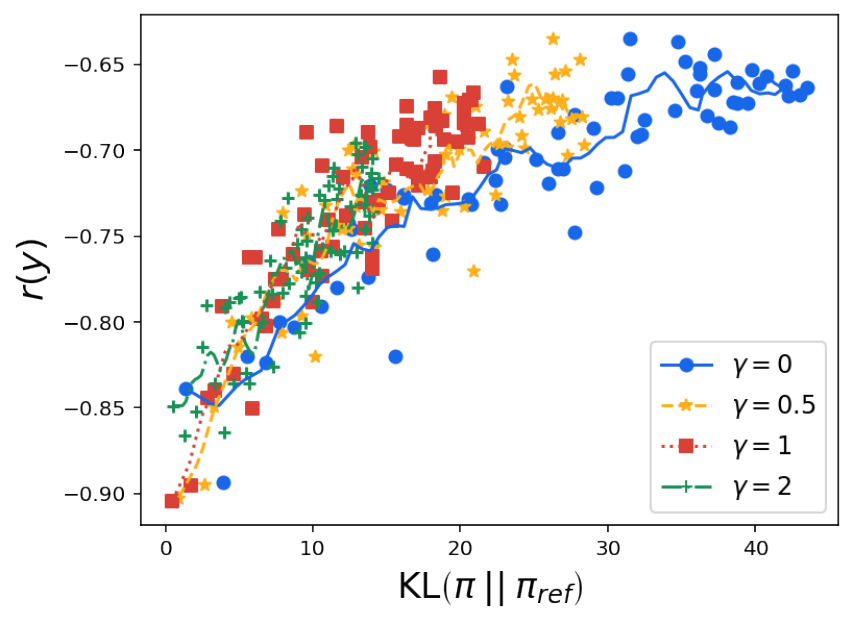}
    \caption{\JBOND: the role of the EMA mixing parameter $\eta$ and regularization $\gamma$. The left plot shows \JBOND with $\gamma=0$, $\eta\in \{0.01, 0.05, 0.1\}$: the larger the $\eta$ the faster the average reward increases. The middle and right plots show \JBOND with $\eta=0.05$ and $\gamma\in \{0, 0.5, 1, 2\}$: the larger the regularization $\gamma$, the slower the policy moves away from $\piref$, improving the reward/KL trade-off.}
    \label{fig:ema_and_gamma}
\end{figure}

\begin{figure}[t]
    \centering
    \includegraphics[width=0.325\textwidth]{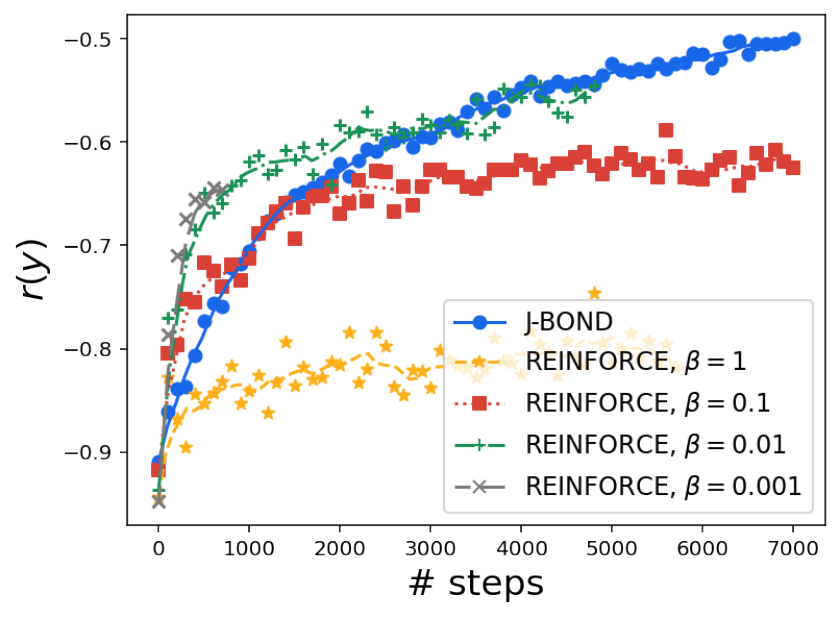}
    \includegraphics[width=0.325\textwidth]{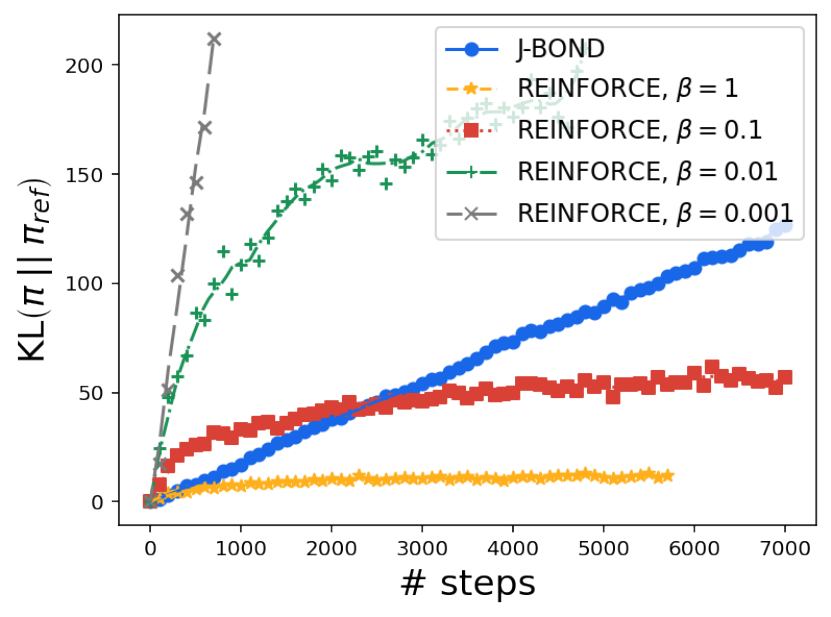}
    \includegraphics[width=0.325\textwidth]{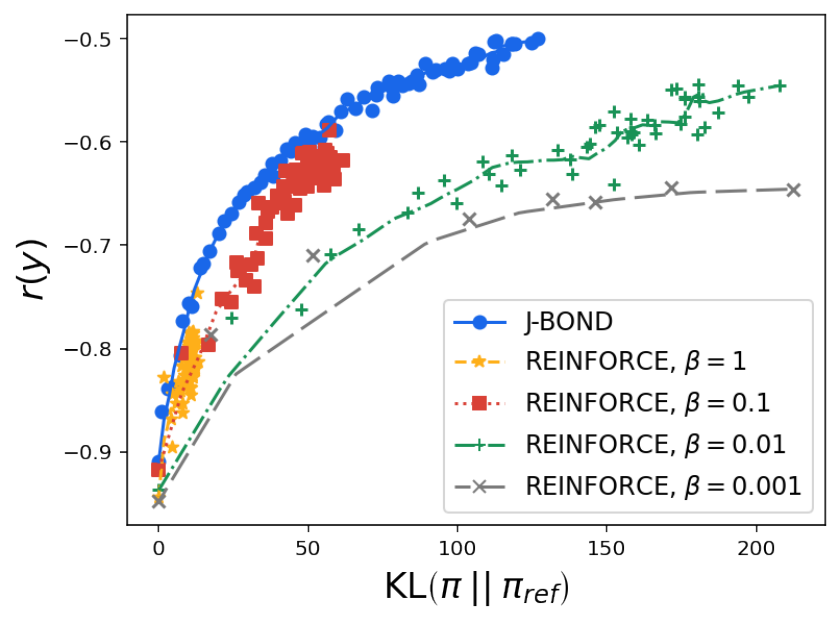}
    \caption{\JBOND ($\eta=0.02$) for Gemma 7B, compared to standard REINFORCE (with KL-regularized objective of \Cref{eq:RL_objective}) with regularization strength $\beta_\textrm{RL} \in \{0.001, 0.01, 0.1, 1\}$. \JBOND does not require committing to a specific regularization strength, but it continuously improves the reward displaying a stable and linear \KL increase and a better reward/\KL trade-off.}
    \label{fig:jbond_vs_reinforce}
\end{figure}

\textbf{Anchor speed and \KL regularization.} 
Here we illustrate the effect of the anchor mixing parameter $\eta \in [0,1]$ and the benefits of the additional \KL regularization parameter $\gamma\geq 0$ introduced in \Cref{eq:j_bond_update_operator}. For this, we consider Gemma 2B and first run \JBOND with $\gamma=0$ and $\eta\in \{0.01, 0.05, 0.1\}$. In the left plot of \Cref{fig:ema_and_gamma} we report the average reward of the policy along training. This illustrates that \emph{the larger the mixing parameter $\eta$} (\ie the faster the anchor moves), \emph{the faster the reward increases}, as one could intuitively expect. Second, we fix $\eta=0.05$ and run \JBOND with different regularization strengths $\gamma\in \{0, 0.5, 1, 2\}$.
We plot the results in the middle and rightmost plots of \Cref{fig:ema_and_gamma}. 
As expected, \emph{the larger the regularization $\gamma$}, the more constrained are the policy updates and thus, \emph{the slower the policy moves away from $\piref$} (middle plot). Importantly, the right plot shows that such a regularization has a positive effect since it can ultimately improve the reward/\KL trade-off.

\textbf{Comparison with standard RLHF.} We compare \JBOND against standard RLHF algorithms that aim at maximizing the \KL-regularized objective of \Cref{eq:RL_objective}. To optimize \Cref{eq:RL_objective}, we use REINFORCE~\citep{williams1992simple} with 2 policy samples per-prompt and a leave-one-out baseline~\citep{ahmadian2024back} for policy gradient advantages. For \JBOND we set the anchor mixing coefficient to $\eta=0.02$. For REINFORCE, we test possible regularization strengths $\beta_\textrm{RL} \in \{0.001, 0.01, 0.1, 1\}$. In \Cref{fig:jbond_vs_reinforce} we plot the average reward of the training policy (left plot) and its KL divergence from the reference $\piref$ (middle plot). As presumed, REINFORCE is quite sensitive to the regularization coefficient $\beta_\textrm{RL}$: the larger the regularization, the lower the reward achieved by REINFORCE (and the lower the KL from $\piref$). This highlight a key advantage of \JBOND: it does not require committing to a specific regularization level, but it continuously improves the reward displaying a stable and linear KL increase. Moreover, in the rightmost plot we plot the corresponding reward/KL trade-off showing that \JBOND produces a better reward/KL than all of the REINFORCE baselines.

\textbf{\JBOND for Gemma open models.} 
\JBOND was also the algorithm used to fine-tune open-weight models such as Gemma 1.1 2B and 7B~\citep{team2024gemma}, RecurrentGemma 2B and 9B~\citep{botev2024recurrentgemmamovingpasttransformers} as well as CodeGemma 1.1~\citep{codegemmateam2024codegemmaopencodemodels}. This led to competitive performance. For example, the Gemma 1.1 IT 7B model outperforms Mistral 7B v0.2 Instruct in both safety and instruction following (see~\citet[Table 5]{team2024gemma} for more details).

\section{Related Work}

\textbf{Best-of-N} was introduced in ~\citet{stiennon2020learning} as a straightforward but costly inference method to optimize language generation against a given reward function. Further works established and refined an analytical form for the KL divergence against the reference (\ie Bo$1$) policy~\citep{hilton2023kl, beirami2024theoretical}, provided an estimator for the average \BON reward~\citep{nakano2021webgpt}, made theoretical connections with \KL-constrained RL~\citep{yang2024asymptotics} and provided scaling laws for \BON alignment~\citep{gao2022scaling}.

\textbf{Matching Best-of-N for improved alignment} is a strategy that was studied in different flavors in the literature. \citet{dong2023raft} and \citet{touvron2023llama2} propose to fine-tune LLMs in a supervised fashion on \BON data actually applying forward \KL minimization. 
Concurrently to ours, \citet{gui2024bonbonalignmentlargelanguage} proposes to mimic the \BON policy by applying a combination of supervised fine-tuning on best responses and direct preference optimization on best-and-worst response pairs.
The latter is similar to a common strategy in online preference optimization methods: \citet{guo2024direct} use pairwise AI feedback on online generations to obtain online preferences that are then optimized, while \citet{calandriello2024human} use a dedicated preference reward model instead. 
Concurrently and closest to our work, \citet{amini2024variationalbestofnalignment} also apply distribution matching in order to get the benefits of \BON sampling with amortized cost. While their formalization is identical, we opt for a different divergence (\ie Jeffreys) than the one they use (\ie only backward KL), and propose an iterative procedure with dynamic anchor, which we show critical for optimal results.
\BON can also be used for self-improvement in reward modeling, as evidenced in~\citet{pace2024west}.

\textbf{Using a contrastive advantage} is an option of \JBOND studied in prior works as well, which replaced a value estimate by the average Monte Carlo return of other samples. This was applied in the context of REINFORCE~\citep{kool2019buy,pinto2023tuning}, for online RLHF~\citep{ahmadian2024back}, offline RLHF~\citep{flet2024contrastive} and preference optimization~\citep{wu2024self}. 

\textbf{Exponential moving average (EMA) of policy as reference} in regularization, which we use in \JBOND, is an increasingly popular option. While most alignment approaches use a static anchor, dynamic anchors bring the benefit of improving the flexibility of the policy space being explored~\citep{munos2023nash, gorbatovski2024learn, rame2024warp}, with the caveat that too slow updates limit optimization and too fast updates hinder stability.

\textbf{Scaling post-training and iterated amplification}.
\BOND hinges on the idea of investing more resources during training to ensure that computational demands during inference remain low, a factor often overlooked in traditional scaling laws \citep{hoffmann2022training}.
Specifically, \BOND incorporates the principles of iterated amplification \citep{christiano2018supervising,cotra2018idamplification}, where amplification in this context consists of producing multiple generations, comparing their rewards, and using these to iteratively improve the policy performance.
In this regard, \BOND is complementary to WARM \citep{rame2024warm} and WARP \citep{rame2024warp}, which previously scaled post-training by training multiple reward models and policies, respectively.

\section{Conclusion}%
\label{sec:conclusion}%
We introduce \BOND, a novel RLHF method that fine-tunes the policy via online distillation of the \BON sampling distribution.
We propose a concrete algorithm, \JBOND, that integrates multiple components to enhance its practicality and efficiency; Monte-Carlo quantile estimation, a combination between forward and backward \KL divergence objectives, and an iterative procedure with an exponential moving average anchor.
\JBOND improves the \KL-reward Pareto front of solutions, and compares favorably against state-of-the-art baselines.
We hope this work can help improve alignment of AI systems, making them safer and more reliable.

\section*{Acknowledgements}

We thank Daniele Calandriello for insightful comments, as well as Gil Shamir, Bilal Piot, and Remi Munos for helpful discussions.
\bibliography{main}
\clearpage
\appendix

\section{Supporting results and derivations}

\subsection{Proof of~\Cref{thm:bondist}}\label{sec:proof_thm}

Consider $N$ random generations $y_1$, $y_2$, \dots, $y_N$ from $\piref$ and an arbitrary generation $y$ among them. 
Let $A_i(y)$ denote the event that $y$ is the best sample (\ie $r(y) \geq r(y_i)$ for all $i$) and that $i$ is the lowest index for which $y_i=y$.
It is trivial to see that the the events $\{A_i(y)\}_{i=1,2, \dots, N}$ are disjoint and that their union corresponds to $y$ being selected by \BON sampling.

The event $A_i(y)$ occurs if and only if three conditions are met: $r(y_j) < r(y)$ for all $j < i$, $y_i =y$, and $r(y_j) < r(y)$ for all $j < i$.
This allows us to derive the likelihood of the event $A_i(y)$:

\begin{align*}
    \mathbb{P}\left[A_i(y)\right] &= \left(\prod_{j=1}^{i-1} \mathbb{P}\left[r(y_j) < r(y)\right]\right)\times \piref(y) \times \left(\prod_{j=i+1}^{N} \mathbb{P}\left[r(y_j) \leq r(y)\right]\right) \\
    &= \pl(y)^{i-1} \times \piref(y) \times \ple(y)^{N-i-1}.
\end{align*}

The likelihood that \BON sampling selects the generation $y$ is then given by
\begin{align*}
    \piBoN(y) &= \sum_{i=1}^N \mathbb{P}\left[A_i(y)\right]\\
    &= \sum_{i=1}^N \left[\pl(y)^{i-1} \times \piref(y) \times \ple(y)^{N-i}\right] \\
    &= \piref(y) \times \sum_{i=1}^N \left[\pl(y)^{i-1} \times  \ple(y)^{N-i}\right] \\
    &= {\piref(y)} \times \ple(y)^{N-1} \times \sum_{i=1}^N \left[\frac{\pl(y)}{\ple(y)}\right]^{i-1}. \hfill \qed
\end{align*}

\subsection{Link to the continuous case}\label{app:continuous_bon}
A noteworthy observation is that we can relate the \BON expression to the case of a continuous distribution, in which case the term $\mathtt{(B)}$ in \Cref{eq:bonproba} is constant and equal to $N$ (which is intuitively natural as $\pl(y)$ and $\ple(y)$ have the same value in this case). 

Indeed, recall that the probability for a sequence $y$ to be drawn from the \BON distribution is
\begin{equation}
     \piBoN(y) 
    = \piref(y) \times \underbrace{\ple(y)^{N-1}}_{\mathtt{(A)}} \times \underbrace{\sum_{i=1}^N \left[\frac{\pl(y)}{\ple(y)}\right]^{i-1}}_{\mathtt{(B)}}.
    \label{eq:bonproba2}
\end{equation}
Here, $y$ is a discrete variable, as it lives in $\llbracket 1;T\rrbracket^L$ where $T$ is the number of tokens and $L$ is the maximum length of a sequence.

Now, we show why \Cref{eq:bonproba2} matches the classic formula for the max of $N$ continuous variables. Formally, let $X$ be a real valued random variable with density $f_X$ and a cumulative distribution function $F_X$. Taking $Y_1, ... Y_N$ i.i.d. variables with the same density, define $X_N=\max \{Y_1, ... Y_N\}$ as the maximum over the $N$ variables. Then, we have that  
\begin{equation}
     F_{X_N}(y) = \mathbb{P}(Y_1 \leq y, \dots Y_N \leq y) = F_X(y) ^ N,
\end{equation}
and thus
\begin{equation}
f_{X_N}(y) = f_X(y) F_X(y) ^ {N-1} N.
\label{eq:bonprobacont}
\end{equation}
In \Cref{eq:bonprobacont}, we recognize the \BON formula in the case where the correction factor $\mathtt{(B)}$ is $N$. For the term $\mathtt{(A)}$,   $F_X(y)$ plays the role of $\ple(y)$, as by definition $F_X(y) = \mathbb{P}(X \leq y)$. Finally, $f_X(y)$ is the density of $X$, which is analogous to the probability $\piref(y)$ in the discrete case.

\subsection{Backward KL and policy gradient equivalence}\label{sec:pg_equivalence}

We formally show the analogy between the gradient of the backward KL divergence of~\Cref{eq:backward_kl_from_bon} and the standard (\eg REINFORCE~\citep{williams1992simple}) policy gradient of a KL-regularized RLHF problem with equivalent reward $r_\texttt{BOND}$ and regularization $\beta_\texttt{BOND}$.

The exact backward KL gradient can be derived as:
\begin{align*}
    \nabla_\pi \operatorname{KL}\left( \pi \mid\mid \piBoN \right) &= \nabla_\pi \mathbb{E}_{y\sim \pi}\left[\log \pi(y) - \log \piBoN(y) \right] \\ 
    & = \nabla_\pi \sum_{y} \pi(y) \left(\log \pi(y) - \log \piBoN(y) \right) \\ 
    & = \sum_{y} \nabla_\pi \pi(y) \left(\log \pi(y) - \log \piBoN(y) \right) + \pi(y)\nabla_\pi \log \pi(y) \\
    & = \sum_{y}  \pi(y) \nabla_\pi \log \pi(y) \left(\log \pi(y) - \log \piBoN(y) \right) + \pi(y) \nabla_\pi \log \pi(y) \\ 
    & = \mathbb{E}_{y\sim \pi }\left[ \nabla_\pi \log \pi(y) \left(\log \pi(y) - \log \piBoN(y) \right)  + \nabla_\pi \log \pi(y) \right] \\
    & = \mathbb{E}_{y\sim \pi }\left[ \nabla_\pi \log \pi(y) \left(\log \pi(y) - \log \piBoN(y) \right) \right].
\end{align*}
Above, we have used the product rule of gradient, the rule $\nabla_\pi \pi(y) = \pi(y) \nabla_\pi \log \pi(y)$ and the fact that $\mathbb{E}_{y\sim \pi} \nabla_\pi \log\pi(y)=0$.

\emph{Equivalence with Policy Gradient RL.} As anticipated, one can verify that descending the above gradient is equivalent -- up to a constant scaling -- to running the RL policy gradient REINFORCE algorithm on the RL objective of Equation~\ref{eq:RL_objective} with $r=r_\texttt{BOND}$ and $\beta_\textrm{RL}=\beta_\texttt{BOND}$. Indeed, we can use the expression for $\piBoN$ to break down the above gradient into:
\begin{align*}
    & \mathbb{E}_{y\sim \pi }\left[ \nabla_\pi \log \pi(y) \left(\log \pi(y) - \log \piBoN(y) \right) \right]  \\
    & = \mathbb{E}_{y\sim \pi }\left[ \nabla_\pi \log \pi(y) \left(\log \pi(y) - \log \piref(y) - (N-1) \log \ple(y) - \log \sum_{i=1}^N \left[\frac{\pl(y)}{\ple(y)}\right]^{i-1} \right) \right] \\ 
    & = \mathbb{E}_{y\sim \pi }\left[ \nabla_\pi \log \pi(y) \left(\log \pi(y) - \log \piref(y) - \frac{r_\texttt{BOND}(y)}{N-1} \right)  \right] \\
    & = - (N-1)  \underbrace{\mathbb{E}_{y\sim \pi }\left[  \nabla_\pi \log \pi(y) \left( r_\texttt{BOND}(y) - \beta_\texttt{BOND} \left(\log \pi(y) - \log \piref(y) \right) \right) \right]}_{\text{gradient used by REINFORCE}}.
\end{align*}

\subsection{Derivation of \texttt{J-BOND} reward}\label{sec:jbond_reward}

Here we provide a theoretical explanation behind the design of the \JBOND reward function discussed in~\Cref{sec:jbond}:
\begin{equation*}
   r_\texttt{J-BOND}(y) =  \begin{cases} -\log(16) & \text{if} \quad r(y) \leq \text{min}\{r(y_1'), r(y_2')\}\\ 0 & \text{otherwise} \end{cases}.
\end{equation*}

Recall that $r_\texttt{J-BOND}(\cdot)$ is meant to approximate the the true reward function $r_{\texttt{BOND}}(y)=\log\ple(\cdot)$ which is unknown since $\ple(\cdot)$ in general requires knowing the reward distribution of $\pi_\text{anchor}^t$ (in \JBOND, we only take 2 samples $y_1', y_2' \sim \pi_\text{anchor}^t$). 

As mentioned in~\Cref{sec:jbond}, we designed $r_\texttt{J-BOND}(\cdot)$ to assign a negative reward only if sample $y$ is worse than both the anchor samples, to mimic the concavity of the log quantile $\log\ple(\cdot)$. In practice, we did not observe gains when rewarding also the intermediate case.   The particular choice of value $-\log(16)$ is motivated by the following main reason.

\begin{figure}[h]
    \centering
    \includegraphics[width=0.5\textwidth]{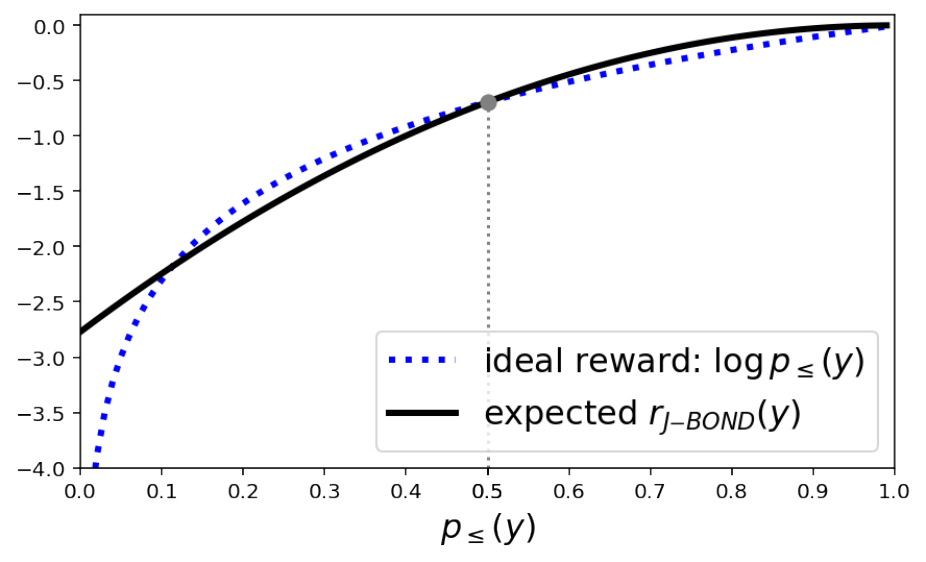}
    \caption{Expected value of the \JBOND reward, \ie $\mathbb{E}_{y_1', y_2' \sim \pi_\text{anchor}^t}\left[ r_\texttt{J-BOND}(y)\right]$ (see~\Cref{eq:expected_jbond_reward}), compared to the true (unknown) reward $\log \ple(y)$, as a function of the quantile $\ple(y)$. By design of $r_\texttt{J-BOND}$, the two curves coincide when $\ple(y) = 0.5$, \ie when $y$ has median reward w.r.t. the anchor distribution.}
    \label{fig:jbond_reward_vs_ideal}
\end{figure}

We want that, when sample $y$ has median reward compared to the anchor rewards' distribution (\ie $\ple(y)=0.5$), then -- in expectation -- $r_\texttt{J-BOND}(y)$ coincides with the true reward $r_{\texttt{BOND}}(y)=\log\ple(\cdot)=\log(0.5)$. For this purpose, let us consider the parametrized function:
\begin{equation*}
   r_\texttt{J-BOND}^\alpha(y) =  \begin{cases} \alpha & \text{if} \quad r(y) < \text{min}\{r(y_1'), r(y_2')\}\\ 0 & \text{otherwise} \end{cases}.
\end{equation*}
Note that the stochasticity of $r_\texttt{J-BOND}^\alpha(y)$ is due to the 2 random anchor samples $y_1', y_2'$ and its expectation can be computed as: 
\begin{align}\label{eq:expected_jbond_reward}
    \mathbb{E}_{y_1', y_2' \sim \pi_\text{anchor}^t}\left[ r_\texttt{J-BOND}^\alpha(y)\right] & = \alpha \cdot \mathbb{P}\left[\{r(y_1') > r(y)\} \cap \{r(y_2') > r(y)\} \right] +  0 \cdot \mathbb{P}\left[ \text{"otherwise"} \right] \\
    & = \alpha \cdot (1-\ple(y))^2,
\end{align}
where we have used the definition of $\ple(y) = \mathbb{P}_{y'\sim\pi_\text{anchor}^t}\left[r(y') \leq r(y)\right]$. Using the expression above, we can find the $\alpha$ for which $\mathbb{E}_{y_1', y_2' \sim \pi_\text{anchor}^t}\left[ r_\texttt{J-BOND}^\alpha(y)\right] = r_\texttt{\BOND}(y)$ when  $\ple(y)=0.5$: 
$$\alpha \cdot(1-0.5)^2 = \log(0.5)\quad \rightarrow \alpha = -\log(16).$$
We illustrate this in~\Cref{fig:jbond_reward_vs_ideal}, where we plot the expected $r_\texttt{J-BOND}(y)$ reward and the true reward $r_\texttt{\BOND}(y)$ as a function of $\ple(\cdot)$.

\section{Additional Experiments}

\subsection{Learned quantile models}\label{app:learned_quantiles}
Monte-Carlo quantile estimation (\Cref{sec:montecarlo}) approximates the reward quantiles by sampling multiple times from the reference policy $\piref$, for each observed context. While we found it to be very simple and effective, it may require many sample for an accurate quantile estimation and, in addition, it does not exploit any information about the given context. For instance, assuming we have a good quantile estimation for context $x$ and are presented a new context $x'$. MC quantile estimates treat $x'$ independently from $x$, although they may have very similar reward quantiles.

Motivated by this, in this section we explore an alternative approach that aims at \emph{learning} a context-dependent quantile estimator $\widehat{\ple}_\theta(\cdot)$, parametrized by parameter $\theta$.
The idea is to view quantile $\ple(y)$ as the parameter of a Binomial random variable $Z$ where $Z=\mathbb{I}\{ r(y_i) \leq r(y)\}$ for $y_i\sim\piref$. Under such a view, we can interpret $\widehat{\ple}_\theta(\cdot)$ as the output of a binary classifier and train it via maximum likelihood estimation using the standard binary cross-entropy loss~\citep{cover1999elements}:
\begin{equation}\label{eq:loss_learned_quantiled}
L(\theta, x, y) = - \mathbb{E}_{y'\sim\piref(x)}\left[\log \widehat{\ple}_\theta(x, y)\mathbb{I}_{\{ r(x,y') \leq r(x,y)\}} + \log(1-\widehat{\ple}_\theta(x,y))\mathbb{I}_{\{ r(x,y') > r(x,y)\}}\right].
\end{equation}

\begin{figure}[h]
    \centering
    \includegraphics[width=0.32\textwidth]{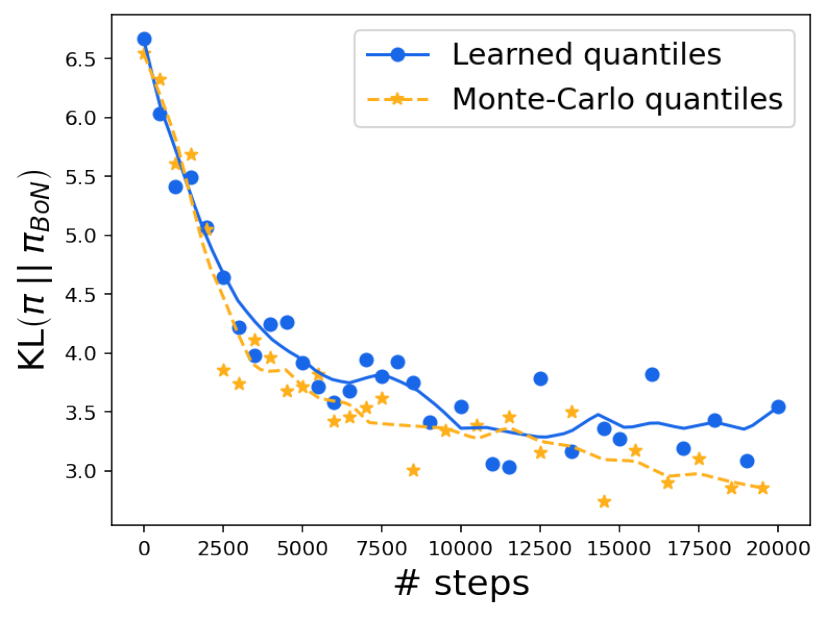}
    \includegraphics[width=0.325\textwidth]{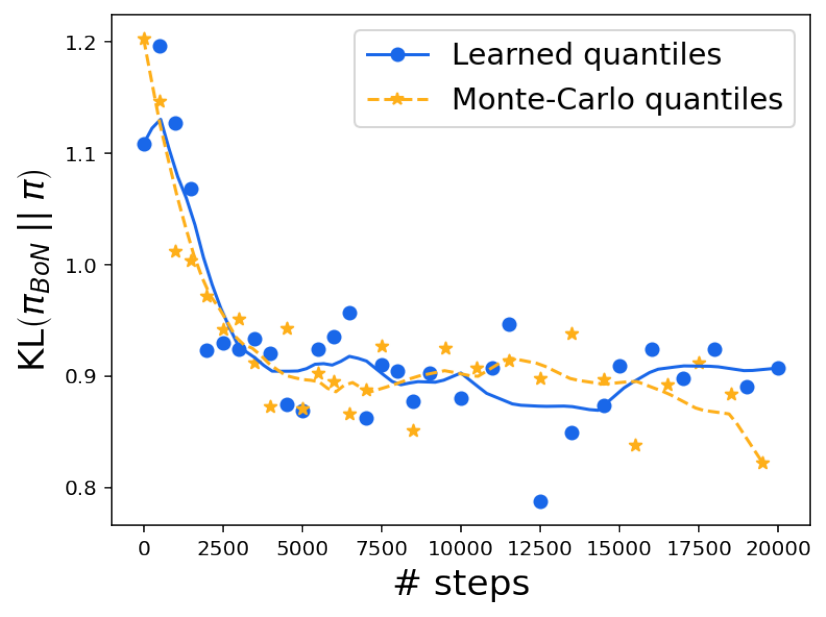}
    \includegraphics[width=0.33\textwidth]{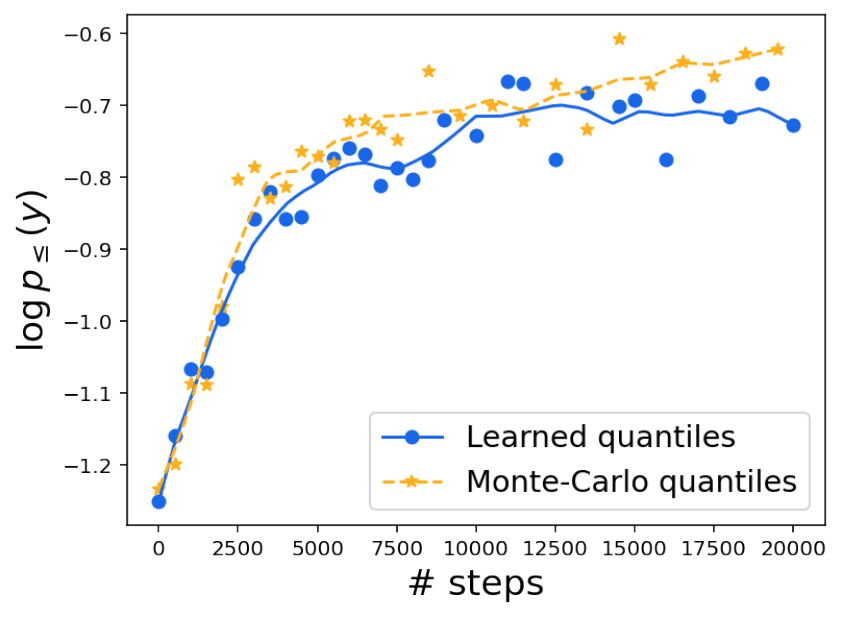}
    \caption{\texttt{BOND} (with $N=8$) using MC quantile estimates vs. a Learned quantile model.}
    \label{fig:learned_quantiles}
\end{figure}

We test such an approach in the abtractive summarization task considered in~\Cref{sec:bond_algorithms}. We parametrize $\widehat{\ple}_\theta(\cdot)$ with a LLM initialized as $\piref$ and fine-tuned using the loss of~\Cref{eq:loss_learned_quantiled}. Simultaneously, the policy $\pi_t$ is fine-tuned using \BOND and utilizing $\widehat{\ple}_{\theta_t}(\cdot)$ as quantile estimator at each step. Notably, we approximate the expectation in~\Cref{eq:loss_learned_quantiled} via a \emph{single} sample from $\piref$ for each prompt in the batch. In~\Cref{fig:learned_quantiles} we report the backward and forward \KL divergences between the training policy and the $\piBoN$ distribution as well as the average log quantiles, and compare them to the ones obtained when running \BOND with MC quantile estimation. Note that in both cases, $\piBoN$ is approximated by 32 MC samples during evaluation.
When using the learned quantile model $\widehat{\ple}_{\theta}(\cdot)$, we observe \BOND achieves very comparable KL divergences and log quantiles compared to using MC quantile estimaton. This illustrates that the use of learned quantiles is valid and promising, potentially offering interesting computational advantages in situations where, \eg $\widehat{\ple}_{\theta}(\cdot)$ can be re-used or learned offline with a fixed sample budget.

Finally, we remark that the explored approach is quite naive, and alternative learned quantile models can definitely be derived, \eg further enforcing ordering in the predicted quantiles, using quantile regression~\citep{Dabney2017DistributionalRL}, or assuming a pre-specified (\eg Gaussian) rewards' distributions.

\subsection{Additional plots for \texttt{BOND} with Jeffreys divergence objective}\label{app:additional_results_jeffreys}
\begin{figure}[h]
    \centering
    \includegraphics[width=0.32\textwidth]{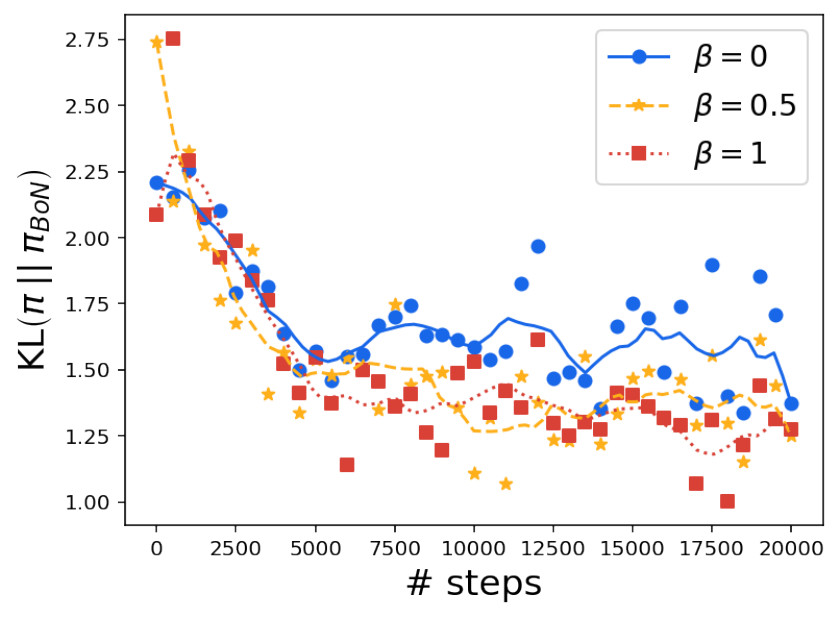}
    \includegraphics[width=0.325\textwidth]{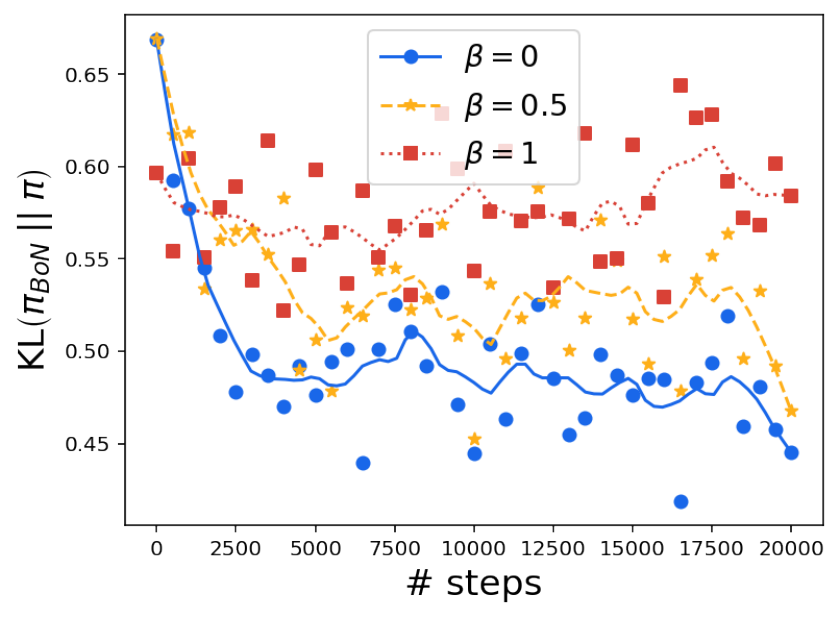}
    \includegraphics[width=0.33\textwidth]{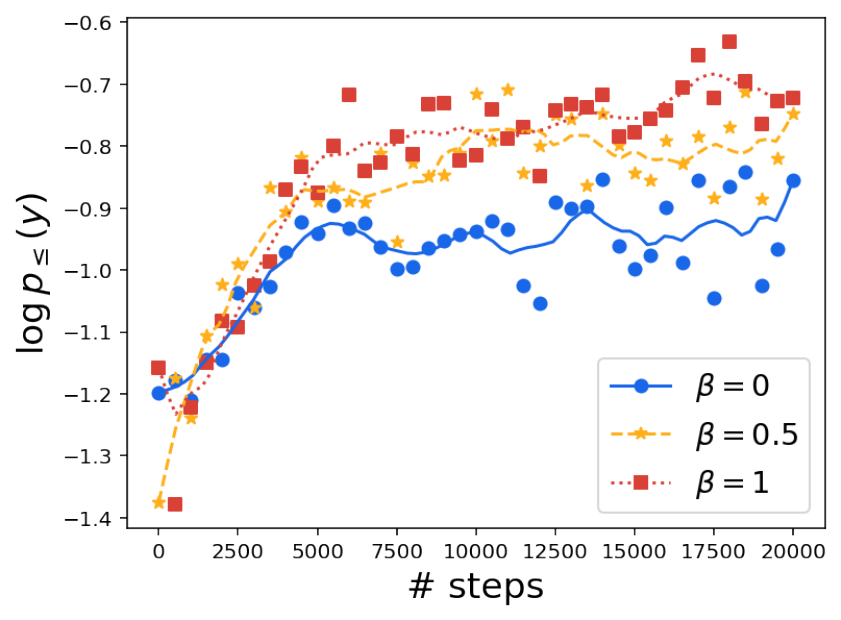}
    \includegraphics[width=0.32\textwidth]{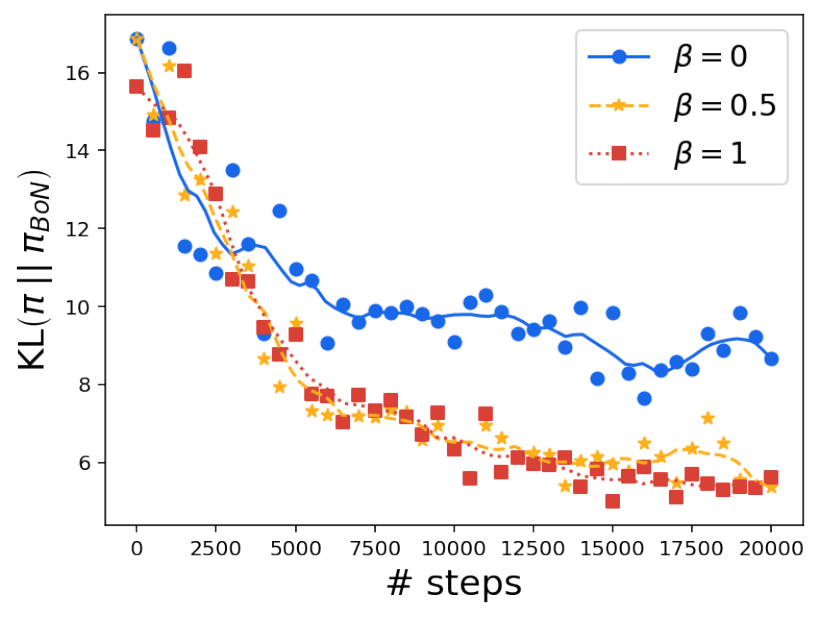}
    \includegraphics[width=0.325\textwidth]{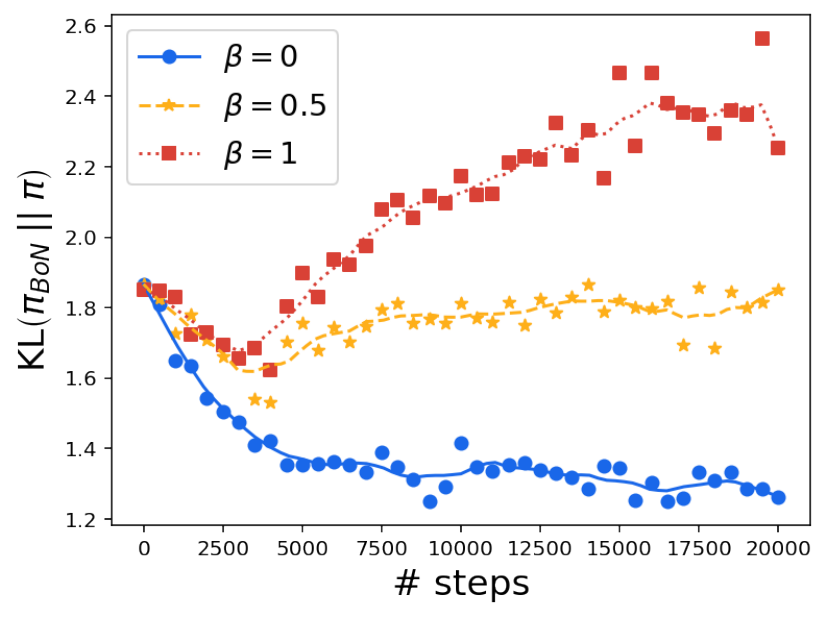}
    \includegraphics[width=0.33\textwidth]{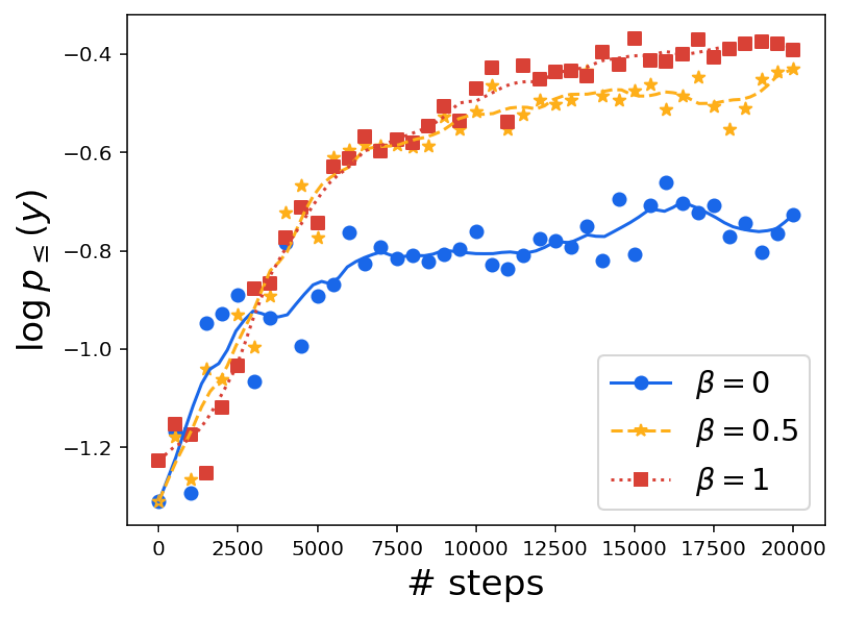}
    \caption{\BOND with $N=4$ (top plots) and $N=16$ (bottom plots), and different values of $\beta$ for the Jeffreys divergence objective (\emph{cf.} \Cref{eq:jeffreys_div}). When using $\beta=0.5$ (Jeffreys divergence), \BOND minimizes both backward (left plots) and forward (middle plots) KL divergences from $\piBoN$, achieving best of both objectives ($\beta=0$ and $\beta=1$). Moreover, it optimizes the reward quantiles (right plots) significantly more than when using $\beta=0$.}
    \label{fig:jeffreys_ablation_N4_and_16}
\end{figure}

We provide additional experiments that complement the ablation of~\Cref{sec:jeffreys} when running \BOND with a Jeffreys divergence objective $J_\text{effreys}^\beta$ for different values of $\beta$. In particular, in~\Cref{fig:jeffreys_ablation_N4_and_16} we show \BOND results when using $N=4$ (top plots) and $N=16$ (bottom plots). Results are consistent with what discussed in~\Cref{sec:jeffreys}: When using $\beta=0.5$, \BOND minimizes both backward and forward KL divergences from $\piBoN$ compared to when solely one of them is used as \BOND objective. Moreover, it optimizes the log quantiles similarly to $\beta=1$.

\end{document}